\documentclass[journal]{IEEEtran}

\ifCLASSINFOpdf
\else
\fi

\usepackage{graphicx}
\usepackage{amsmath,amssymb} % define this before the line numbering.
\usepackage{color}
\usepackage{subfigure}
\usepackage{verbatim}
\usepackage{cite}
\usepackage{url}
\usepackage{times}
\usepackage{epsfig}
\usepackage{multirow}
\usepackage{color}
\usepackage{cite}

\def\ie{\emph{i.e.~}}
\def\eg{\emph{e.g.~}}
\def\etal{{\em et al.~}}
\newcommand{\degree}{^\circ}

\begin{document}

\title{How is Gaze Influenced by Image \\ Transformations? Dataset and Model}

\author{Zhaohui~Che,
        Ali~Borji,~\IEEEmembership{Member,~IEEE,}
        Guangtao~Zhai,~\IEEEmembership{Senior Member,~IEEE,}
        Xiongkuo~Min,~\IEEEmembership{Member,~IEEE,}
        Guodong~Guo,~\IEEEmembership{~Senior~Member,~IEEE}
        ~and~Patrick~Le~Callet,~\IEEEmembership{~Fellow,~IEEE}
\thanks{Manuscript received March 04, 2019; revised August 29, 2019; accepted September 26, 2019. Date of publication XXXX-XXXX, XXXX; date of current version XXXX-XXXX, XXXX.  This work was supported by the National Science Foundation of China (61831015, 61771305, 61927809 and 61901260), and in part by the China Postdoctoral Science Foundation under Grants BX20180197 and 2019M651496. The associate editor coordinating the review of this manuscript and approving it for publication was Prof. Soma Biswas. (\textit{Corresponding author: Guangtao Zhai.})}

\thanks{Z. Che, G. Zhai, X. Min are with the Institute of Image Communication and Network Engineering, Shanghai
Key Laboratory of Digital Media Processing and Transmissions, Shanghai Jiao Tong University, Shanghai 200240, China
(e-mail: \{chezhaohui,zhaiguangtao,minxiongkuo\}@sjtu.edu.cn).}

\thanks{A. Borji is a senior research scientist at MarkableAI Inc, Brooklyn, NY 11201, USA (e-mail: aliborji@gmail.com).}

\thanks{G. Guo is with the Institute of Deep Learning, Baidu Research, Beijing 100193, China, and also with the Department of Computer Science and Electrical Engineering, West Virginia University, Morgantown, WV USA (e-mail: guodong.guo@mail.wvu.edu).}

\thanks{P. L. Callet is with the {\'E}quipe Image, Perception et Interaction, Laboratoire
des Sciences du Num{\'e}rique de Nantes, Universit{\'e} de Nantes, France (e-mail: patrick.lecallet@univ-nantes.fr).}

\thanks{Digital Object Identifier XXXXXXXX}}

\markboth{IEEE Transactions on Image Processing}%
{Z. Che \MakeLowercase{\textit{et al.}}: How is Gaze Influenced by Image Transformations? Dataset and Model}


\maketitle

\begin{abstract}
Data size is the bottleneck for developing deep saliency models, because collecting eye-movement data is very time-consuming and expensive.
Most of current studies on human attention and saliency modeling have used high-quality stereotype stimuli. In real world, however, captured images undergo various types of transformations. Can we use these transformations to augment existing saliency datasets?
Here, we first create a novel saliency dataset including fixations of 10 observers over 1900 images degraded by 19 types of transformations. Second, by analyzing eye movements, we find that observers look at different locations over transformed versus original images. Third, we %investigate the qualification of different distortions when serving as
utilize the new data over transformed images, called data augmentation transformation (DAT), to train deep saliency models. We find that label-preserving DATs with negligible impact on human gaze boost saliency prediction, whereas some other DATs that severely impact human gaze degrade the performance. These label-preserving valid augmentation
transformations provide a solution to enlarge existing saliency datasets. % due to that collecting eye-movement data is very expensive and time-consuming.
Finally, we introduce a novel saliency model based on generative adversarial networks (dubbed GazeGAN). A modified U-Net is utilized as the generator of the GazeGAN, which combines classic ``skip connection'' with a novel ``center-surround connection'' (CSC) module. Our proposed CSC module mitigates trivial artifacts while emphasizing semantic salient regions, and increases model nonlinearity, thus demonstrating better robustness against transformations.
% We also propose a histogram loss based on Alternative Chi-Square Distance (ACS HistLoss) to refine the saliency map in terms of histogram distribution.
Extensive experiments and comparisons indicate that GazeGAN achieves state-of-the-art performance over multiple datasets. We also provide a comprehensive comparison of 22 saliency models on various transformed scenes, which contributes a new robustness benchmark to saliency community. Our code and dataset are available at: \url{https://github.com/CZHQuality/Sal-CFS-GAN}.
\end{abstract}


% \fi
\IEEEpeerreviewmaketitle

\begin{figure}%[!htb]
%\vspace{-0.1cm}
\centering
\subfigure[Original]{\label{fig:edge-a}\includegraphics[height=0.26\linewidth]{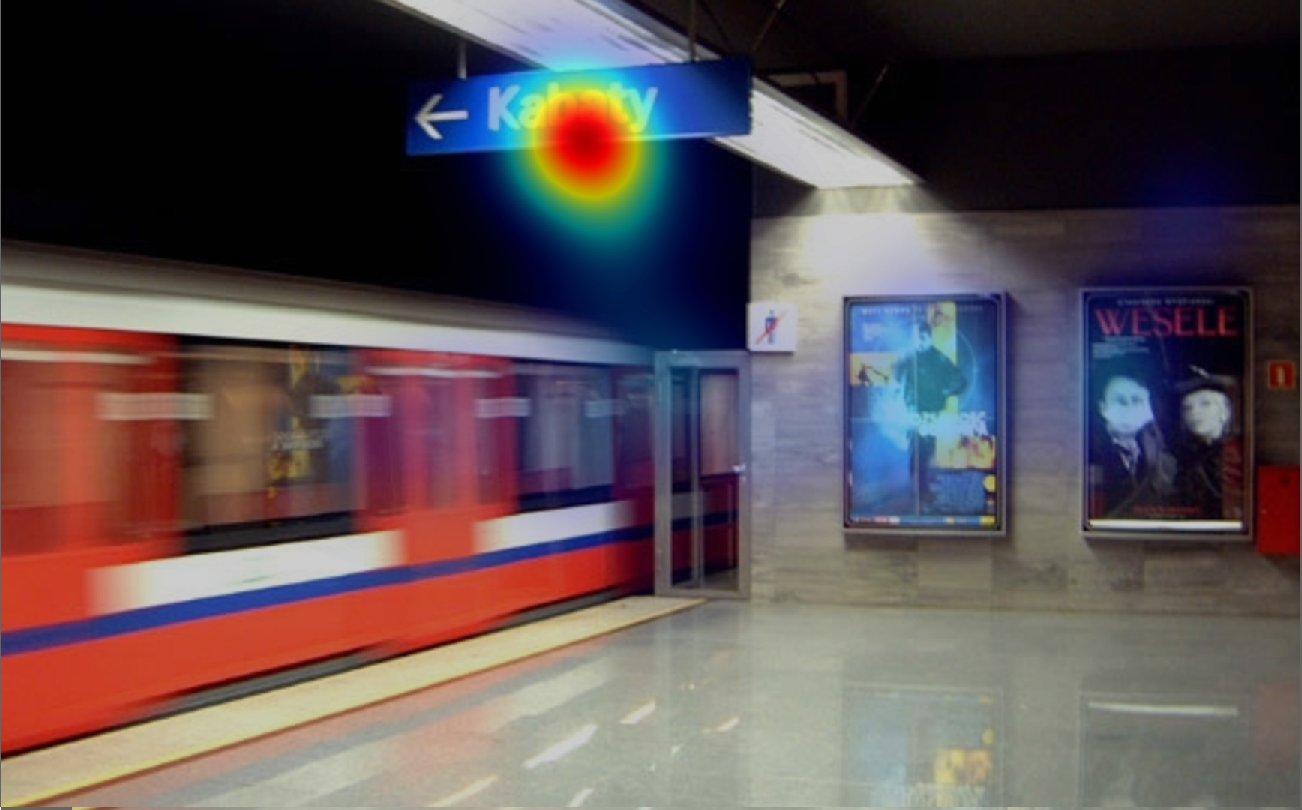}} \hspace{3pt}
\subfigure[Ground Truth]{\label{fig:edge-a}\includegraphics[height=0.26\linewidth]{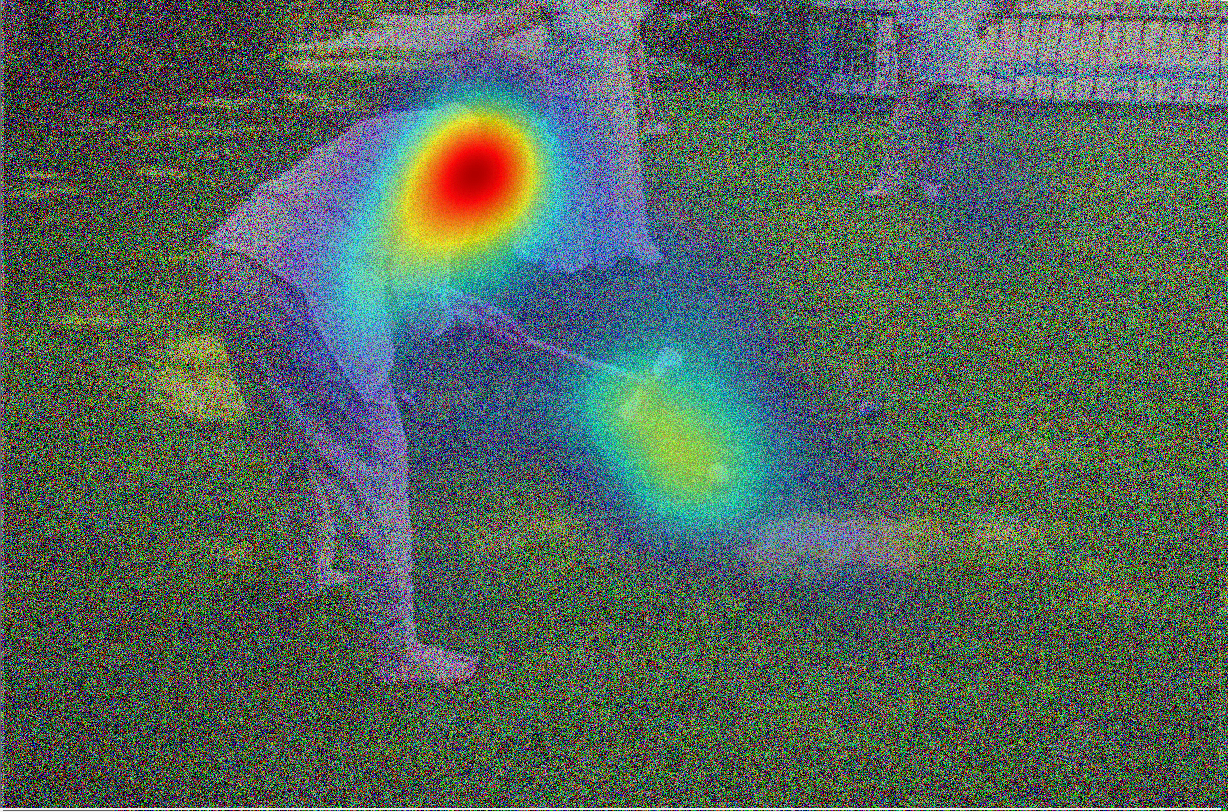}}\\\vspace{-3pt}
\subfigure[Noise]{\label{fig:edge-a}\includegraphics[height=0.26\linewidth]{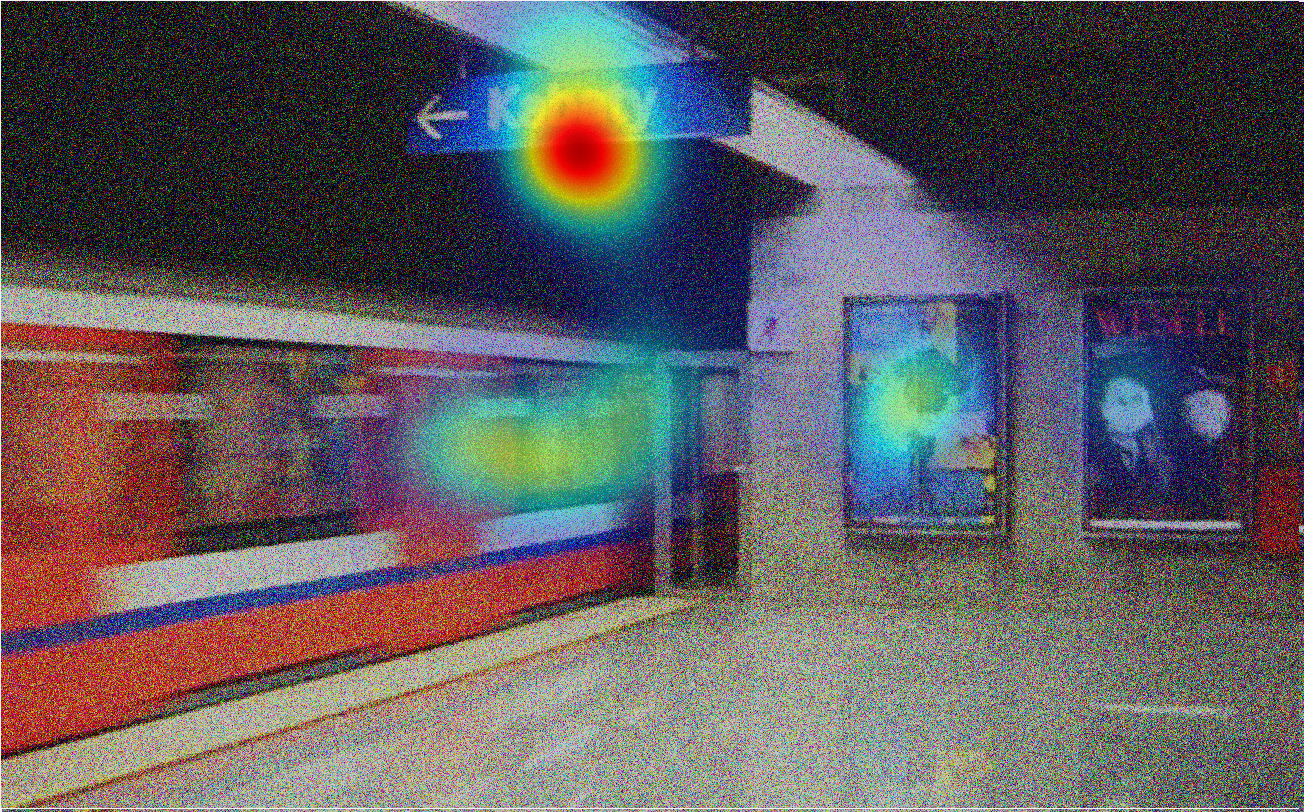}} \hspace{3pt}
\subfigure[SALICON Model]{\label{fig:edge-a}\includegraphics[height=0.26\linewidth]{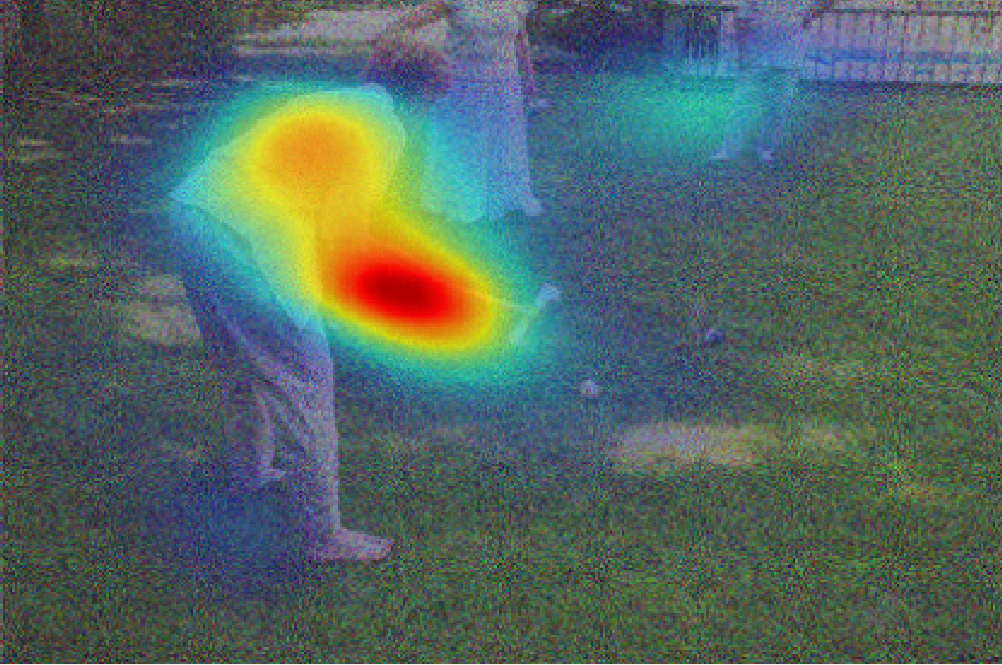}}\\\vspace{-3pt}
\subfigure[Cropping]{\label{fig:edge-a}\includegraphics[height=0.26\linewidth]{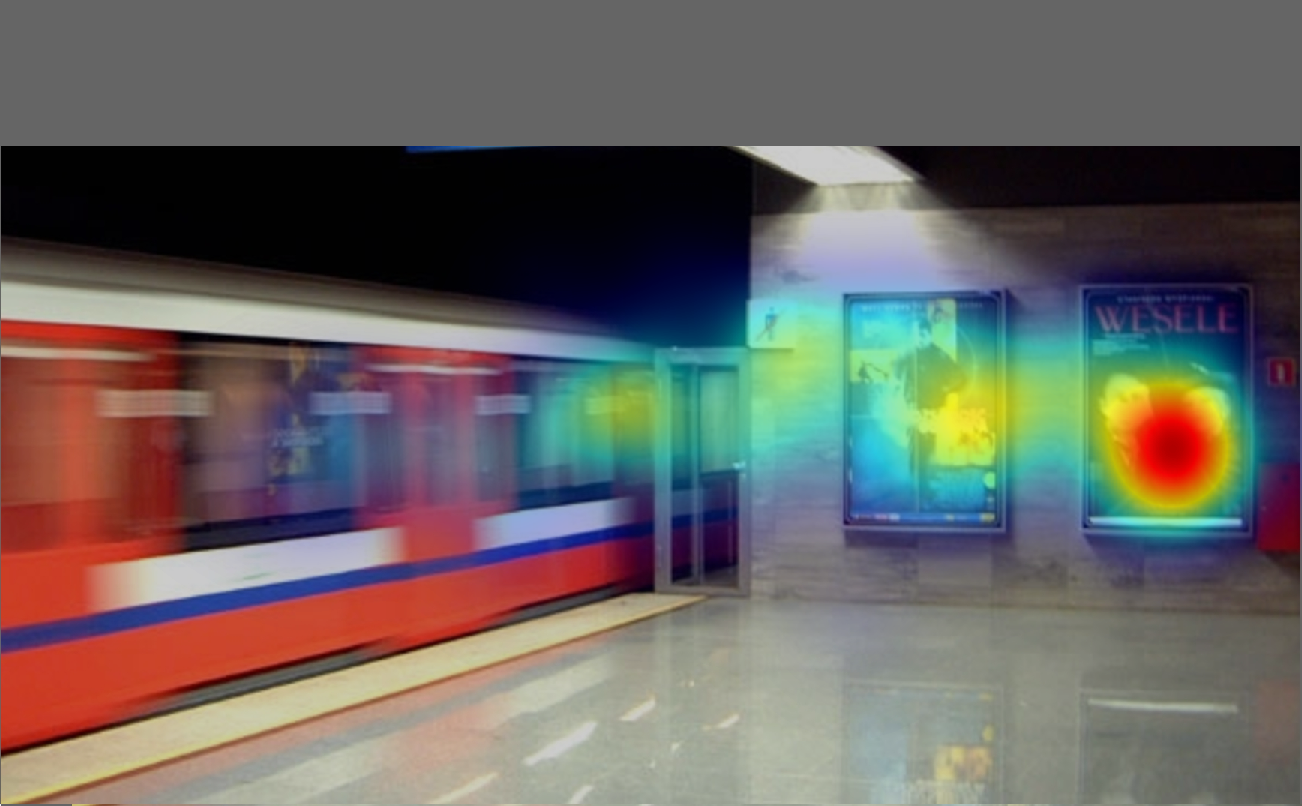}} \hspace{3pt}
\subfigure[Our Model]{\label{fig:edge-a}\includegraphics[height=0.26\linewidth]{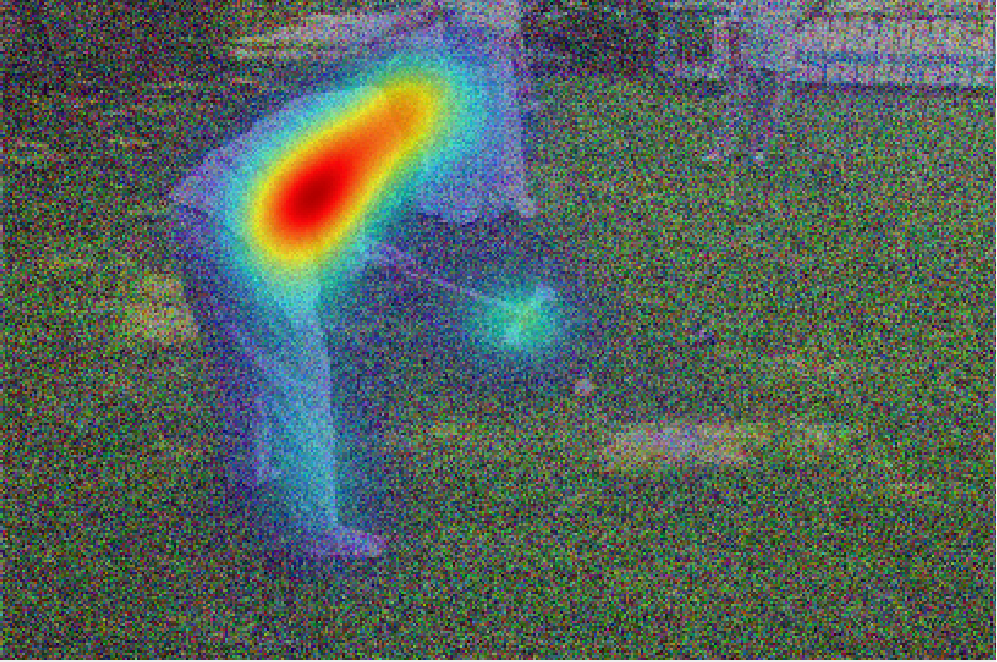}}\\\vspace{-3pt}

\caption{The $1_{st}$ column: (a), (c) and (e) are heatmaps of human gaze on original image, and two transformed versions corrupted by Noise and Cropping. Noise has a slight impact on human gaze, whereas Cropping changes human attention severely. The 2nd column: Prediction results of two saliency models on noisy image. SALICON misses the true positives \ie ``face'', but detects the false positives \eg ``hand''.}
\label{beginfigure}
\end{figure}

\section{Introduction}
\IEEEPARstart{V}{isual} attention is a sophisticated mechanism for selecting informative and conspicuous regions from external stimuli~\cite{borji2013state}. To the best of our knowledge, most of current human attention studies and saliency models are based on stereotype stimuli, \eg~distortion-free images and upright scenes.
However, most of stimuli in the real physical world are corrupted by diverse transformations.

As an example, we present a practical case in the first column of Fig.~\ref{beginfigure}. When viewing the original canonical image, human attention is highly attracted to the ``station board'', because this region provides critical semantic information that helps observers to recognize the scene as a ``railway station''.
On the other hand, when adding noise to this scene, the ``station board'' region still attracts most of human attention.
However, when the ``station board'' is cropped, human gaze is significantly changed. We can see that most of human attention transfers to the ``advertisement board'' and the blurred ``metro'', because these salient objects help observers to understand the new transformed scene. These cases raise new concerns about human gaze invariance on transformed scenes.

In the past decades, a plethora of saliency models \cite{MLnet,SalGAN,SALICON,SalNet,SAM,borji2013state,borji2013quantitative,Borji_2013_ICCV,IttiKoch,GBVS,Torralba,Covsal,AIM,ImageSig,LSLGS,BMS,RC,Murray,AWS,ContextAware,borji2019saliency} have been proposed to detect saliency regions, which serve as an efficient front-end process to complex vision tasks such as scene understanding and object recognition \cite{SalObjDec01,SalObjDec02,SalSegment03}.

Despite their great successes in stereotype clean stimulus, most of current saliency models, either recent deep models or early hand-crafted models, are vulnerable to transformations. As shown in the second column of Fig.~\ref{beginfigure}, the SALICON \cite{SALICON} model is susceptible to noise artifacts, and produces severe false positives such as ``hand'', also misses important true positives like ``face''. Therefore, it is important to investigate new robust approaches to reach the human level accuracy on transformed scenes.

Some related works regard human attention over transformed conditions.
Kim~\etal~\cite{MilanfaSN} investigated visual saliency over noisy images and proposed a model for noise-corrupted images. They found that noise significantly degrades the accuracy of saliency models. Judd~\etal~\cite{JuddLowRes} elaborately investigated gaze over low-resolution images, and compared gaze dispersion on different image resolutions.
Zhang~\etal \cite{ZhangQA} investigated the optimal strategy to integrate attention cues into perceptual quality assessment, and showed that eye-tracking data on transformed images improves perceptual quality assessment methods.

These works, however, only considered certain types of transformations, limited amount of data, and a small set of saliency models. Further, they did not investigate the potential of various transformations for boosting saliency modeling (\eg by serving as data augmentation).
In this paper, we conduct a comprehensive study on the impacts of several transformations on both human gaze and saliency models. We also explore potential application and introduce a robust saliency model.

\section{The Proposed Eye-Movement Database}

\begin{table}
\renewcommand{\arraystretch}{1.5}
\renewcommand{\tabcolsep}{.3mm}

    \caption{ Details of transformations. We list IO scores \cite{LSLGS}, which provide the upper-bound on performance of saliency models.\protect\footnotemark}
    \label{tab:DisGrou}
    \scriptsize
    \begin{tabular}{|c|c|c|}

    \hline
    {\bfseries Transformations} &{\bfseries Generation code (using Matlab)} &{\bfseries {IO scores: sAUC} } \\
   \hline
   \hline
    {\emph{\bfseries Reference}} &{\emph{ 100 distortion-free images (\rm img) from CAT2000}}  &{ {0.733}}\\
    \hline
       \hline
    {{\bfseries MotionBlur1}}  &{{ imfilter(img, fspecial(`motion', \bfseries {15}, \bfseries{0}))}} &{ {0.664}}\\
%    \hline
    {{\bfseries MotionBlur2}}  &{{ imfilter(img, fspecial(`motion', \bfseries{35}, \bfseries{90}))}} &{ \textcolor{black}{0.651}}\\
%    \hline
    {{\bfseries Noise1}}  &{{ imnoise(img, `gaussian', 0, \bfseries{0.1})}} &{ \textcolor{black}{0.706}}\\
    {{\bfseries Noise2}}  &{{ imnoise(img, `gaussian', 0, \bfseries{0.2})}} &{ \textcolor{black}{0.696}}\\
    \hline
       \hline
    {{\bfseries JPEG1}}  &{{ imwrite(img, saveroutine, `Quality', \bfseries{5})}} &{ \textcolor{black}{0.703}}\\
    %\hline
    {{\bfseries JPEG2}}  &{{ imwrite(img, saveroutine, `Quality', \bfseries{0})}} &{ \textcolor{black}{0.705}}\\
    \hline
       \hline
    {{\bfseries Contrast1}}  &{{ imadjust(img, [ ], \bfseries{[0.3, 0.7]})}} &{ \textcolor{black}{0.722}}\\
    %\hline
    {{\bfseries Contrast2}}  &{{ imadjust(img, [ ], \bfseries{[0.4, 0.6]})}} &{ \textcolor{black}{0.702}}\\
    %\hline
    {{\bfseries Rotation1}}  &{{ imrotate(img, \bfseries{-45}, `bilinear', `loose') }} &{ \textcolor{black}{0.680}}\\
    %\hline
    {{\bfseries Rotation2}}  &{{ imrotate(img, \bfseries{-135}, `bilinear', `loose') }} &{ \textcolor{black}{0.654}}\\
    %\hline
    {{\bfseries Shearing1}}  &{{ imwarp(img, \bfseries affine2d([1 0 0; 0.5 1 0; 0 0 1]) }} &{ \textcolor{black}{0.711}}\\
    %\hline
    {{\bfseries Shearing2}}  &{{ imwarp(img, \bfseries affine2d([1 0.5 0; 0 1 0; 0 0 1]) }} &{ \textcolor{black}{0.687}}\\
    %\hline
    {{\bfseries Shearing3}}  &{{ imwarp(img, \bfseries affine2d([1 0.5 0; 0.5 1 0; 0 0 1]) }} &{ \textcolor{black}{0.665}}\\
    \hline
    \hline
    {{\bfseries Inversion}}  &{{ imrotate(img, \bfseries{-180}, `bilinear', `loose') }} &{ \textcolor{black}{0.695}}\\
    %\hline
    {{\bfseries Mirroring}}  &{{ mirror symmetry version of reference images}} &{ \textcolor{black}{0.726}}\\
    %\hline
    {{\bfseries Boundary}}  &{{ edge(img, `canny', 0.3, sqrt(2)) }} &{ \textcolor{black}{0.667}}\\
    %\hline
    {{\bfseries Cropping1}}  &{{ a $1080\times200$ band from the \textbf{left} of img}} &{ \textcolor{black}{0.697}}\\
    %\hline
    {{\bfseries Cropping2}}  &{{ a $200\times1920$ band from the \textbf{top side} of img }} &{ \textcolor{black}{0.692}}\\

     \hline
   %\hline
    \end{tabular}
%\vspace{-15pt}
\end{table}
{\footnotetext{Please see the supplement for more results on IO scores using CC and NSS metrics.}

\subsection{Stimuli and transformation types}

We selected 100 distortion-free reference images from the \textbf{CAT2000} eye-movement database \cite{CAT2000} since it covers various scenes such as
indoor and outdoor scenes, natural and man-made scenes, synthetic patterns, fractals, and cartoon images.
Considering that different reference
images have different aspect ratios, we padded each image by adding two gray bands to the left and right sides and adjusted the image scale to make sure all images have the same resolution ($1080 \times 1920$).

To systematically assess the influence of ubiquitous transformations on human attention behavior, we choose 19 common transformations that could occur during the whole \textit{image acquisition},
 \textit{transmission}, and \textit{displaying} chain, including:
\begin{itemize}
    \item {\bf Acquisition:} 2 levels of motion blur and 2 levels of Gaussian noise,
    %\vspace{-5pt}
    \item {\bf Transmission:} 2 levels of JPEG compression,
    %\vspace{-5pt}
    \item {\bf Displaying:} 2 levels of contrast change, 2 rotation degrees, and 3 shearing transformations,
    %\vspace{-5pt}
    \item {\bf Other:} inversion, mirroring, line drawing (boundary maps), and 2 types of cropping distortions (to explore gaze variations under extremely abnormal conditions).
\end{itemize}

Eventually, we derive 18 transformed images for each reference image, and a total of 1900 images (18 $\times$ 100 + 100 reference images). Details of transformation types and generation code are shown in Table \ref{tab:DisGrou}. Notably, these transformations are wildly used as data augmentation transformations for training deep neural networks to mitigate overfitting~\cite{DA1}.

\begin{figure*}[!htb]
%\vspace{-0.1cm}
\centering
\subfigure[CC$\uparrow$ similarity matrix]{\label{fig:edge-a}\includegraphics[height=0.33\linewidth]{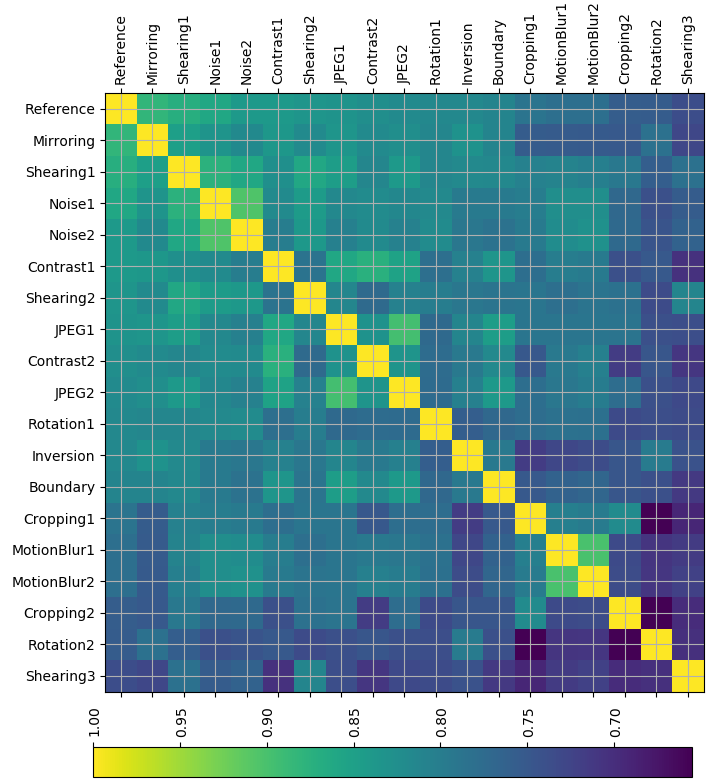}}
\subfigure[SIM$\uparrow$ similarity matrix]{\label{fig:edge-b}\includegraphics[height=0.33\linewidth]{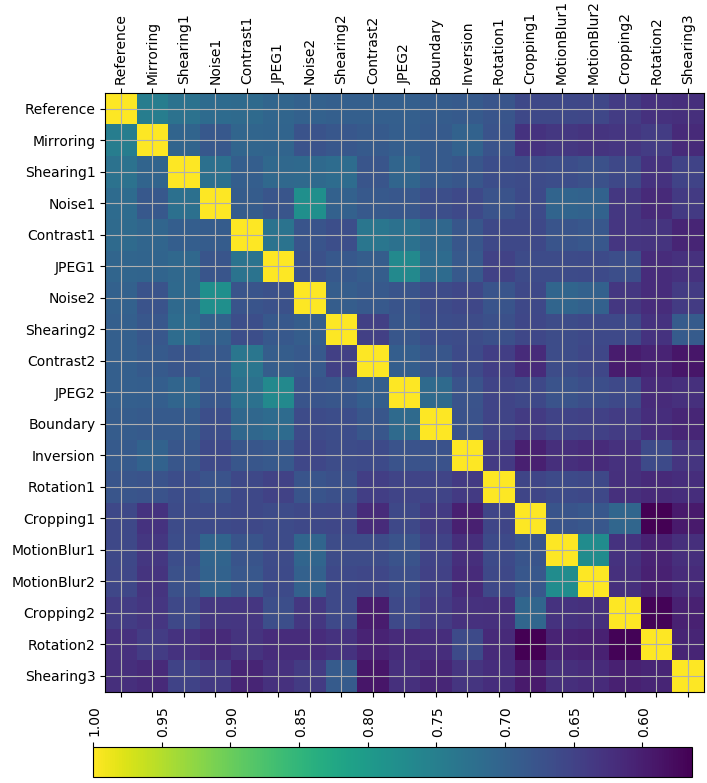}}
\subfigure[KL$\downarrow$ dissimilarity matrix]{\label{fig:edge-c}\includegraphics[height=0.33\linewidth]{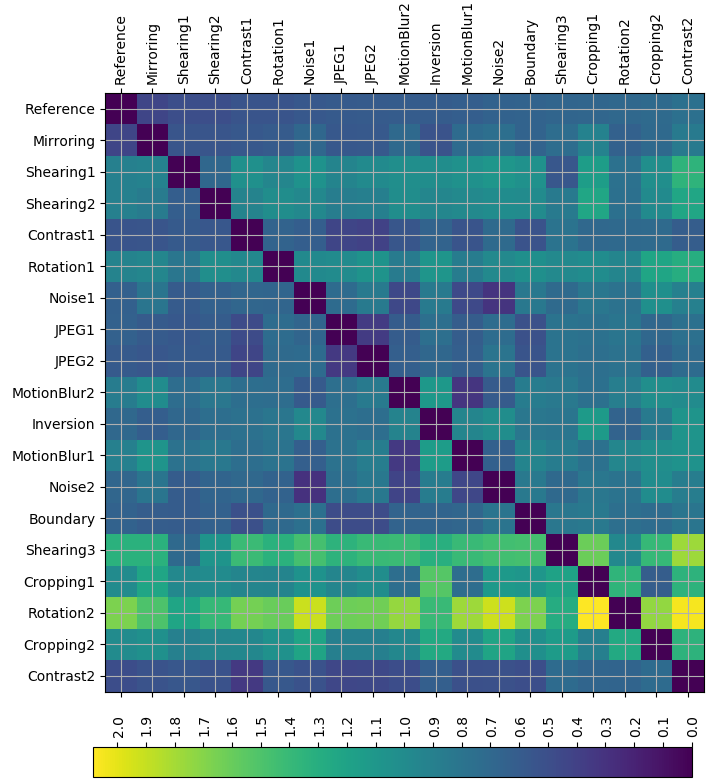}}
\caption{ We plot three similarity/dissimilarity matrices of human gaze when viewing different transformed stimuli versus Reference.
The transformation types are ranked by their similarity/dissimilarity scores when using the human gaze on Reference as ground-truth. The higher CC and SIM values represent the better similarity, while the lower KL value means the better relevance. CC and SIM are symmetric measures, while KL is a non-symmetric measure.}
\label{SMhuman}
\end{figure*}

\subsection{Eye-tracking setup}

As indicated by Bylinskii~\etal \cite{Bylinskii}, the eye-tracking experimental parameters (\eg observers' distance to screen, calibration
error, image size) impact human gaze invariance. To mitigate these issues, we utilized the \textbf{Tobii X120} eye tracker to record eye-movements.
We used the \textbf{LG 47LA6600 CA} monitor with horizontal resolution of 1920 and vertical resolution
 of 1080, to match the resolutions of stimuli and the monitor screen. The height and width of the monitor were 60cm and 106cm, respectively.
 The distance between subject
  and the eye-tracker was 60cm. According to Bylinskii~\etal \cite{VisualAngle}, one degree of visual angle was used both as
  1) an estimate of the size of the human fovea, and
  2) to account for measurement error. In our experiment, the width of the screen subtended $32.81\degree$ of visual angle,
  and $1\degree$ of horizontal angle corresponding to 56.91 pixels ($18.92\degree$ and 56.55 pixels for the screen height, correspondingly).

  Two types of ground-truth data have been traditionally used for training and measuring the accuracy of saliency models:
   1) binary fixation maps made up of discrete gaze points recorded by an eye-tracker, and
   2) continuous density maps representing the probability of the human gaze. The former can be converted into the latter by a Gaussian smoothing filter with
   standard deviation $\sigma$ equal to one degree of visual angle \cite{LeMeur2013}, hence we chose $\sigma=57$ in this paper.

We recruited 40 subjects to participate in the eye tracking experiment under the free-viewing condition. All participants had not been exposed to the stimuli set before. The duration time for each stimulus was $4s$. We inserted a gray image with $1s$ duration between each two consecutive images to reset gaze to the image center for reducing the impact of memory effects \cite{WithinSubject} on gaze invariance. Besides, the presentation order of stimuli was randomized for each subject to mitigate the carryover effect from the previous images.

\section{Analysis of Human Gaze Invariance}

In this section, we quantify the discrepancies between human gaze over transformed and reference images using Pearson's Linear Correlation Coefficient (CC), Histogram Intersection Measure (SIM), and Kullback-Leibler divergence (KL) metrics \cite{mit-saliency-benchmark}.
The CC/SIM similarity matrices and KL dissimilarity matrix are shown in Fig.~\ref{SMhuman}, where the transformation types are ranked by their similarity/dissimilarity values compared to the Reference images.
Since Inversion, Mirroring, Rotation and Shearing transformations change the locations of pixels, we align gaze maps of these transformations with the Reference gaze map via the corresponding inverse transformations for fair comparison.

We first analyse human gaze invariance from a statistical perspective. As shown in Fig.~\ref{SMhuman}, quantitative comparisons on CC, SIM and KL metrics indicate that most of the transformations impact human gaze, and the magnitude of impact highly depends on the transformation type. Besides, different magnitudes of the same transformation have similar impacts on human gaze, \eg Noise1 \emph{vs} Noise2, JPEG1 \emph{vs.} JPEG2, and MotionBlur1 \emph{vs.} MotionBlur2, and higher distortion magnitude causes severer impact. Third, we cannot directly use all of these transformations as data augmentation transformations for saliency prediction, because some transformations are not label-preserving in terms of human gaze.

Next, we provide a fine-grained analysis of human gaze under different transformations from a qualitative perspective.

\begin{figure}
\centering
\subfigure{\label{fig:edge-a}\includegraphics[height=0.48\linewidth]{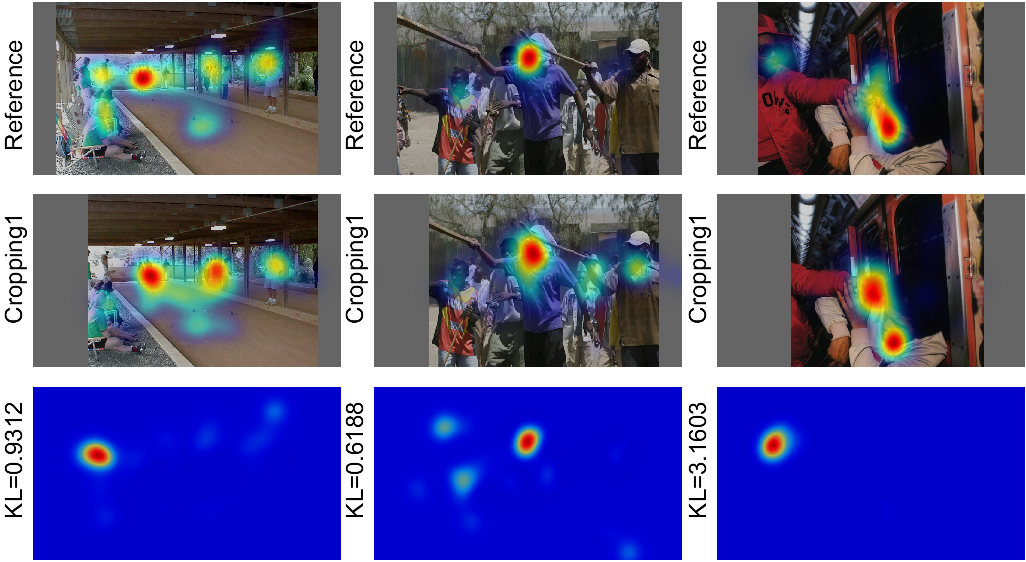}}
\vspace{-0.1cm}
\vspace{-0.1cm}
\caption{ Human gaze discrepancy on Cropping1 compared to Reference. The $1_{st}$ and $2_{nd}$ rows represent the human gaze maps of Reference and Cropping1, respectively. The $3_{rd}$ row represents KL heatmap that highlights discrepant regions, especially the ``lacked'' salient object compared to the Reference image.}
\label{Cropping1}
\end{figure}

\begin{figure}
\centering
\subfigure{\label{fig:edge-a}\includegraphics[height=0.48\linewidth]{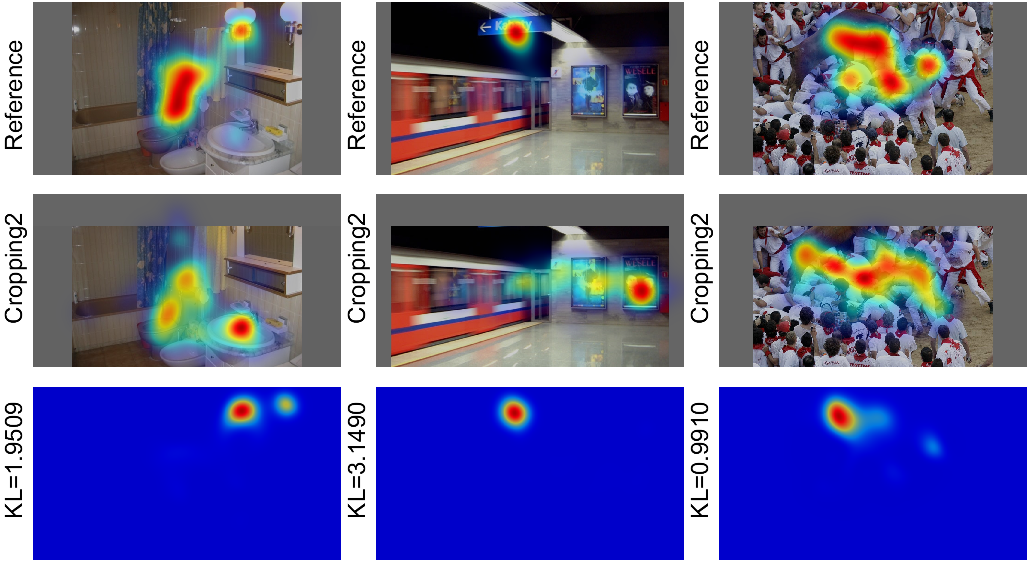}}
\vspace{-0.1cm}
\vspace{-0.3cm}
\caption{ Human gaze discrepancy on Cropping2 compared to the Reference.}
\label{Cropping2}
\end{figure}

\textbf{Cropping:} As shown in Fig.~\ref{Cropping1} and Fig.~\ref{Cropping2}, Cropping transformation may delete some saliency information from the cropped side. For example, in the $2_{nd}$ column of Fig.~\ref{Cropping2}, human attention transfers from ``station board'' to ``advertising boards''. Despite the critical semantic information (\ie ``station board'') being cropped, observers can still recognize the cropped image as a ``railway station'' via new salient objects (\ie ``advertising boards'' and ``metro''). Thus, we arrive at the following empirical inference. \textit{When a scene is cropped, human gaze tends to focus on salient regions with more semantic information that help understand the cropped scene.}

\begin{figure}
\centering
\subfigure{\label{fig:edge-a}\includegraphics[height=0.8\linewidth]{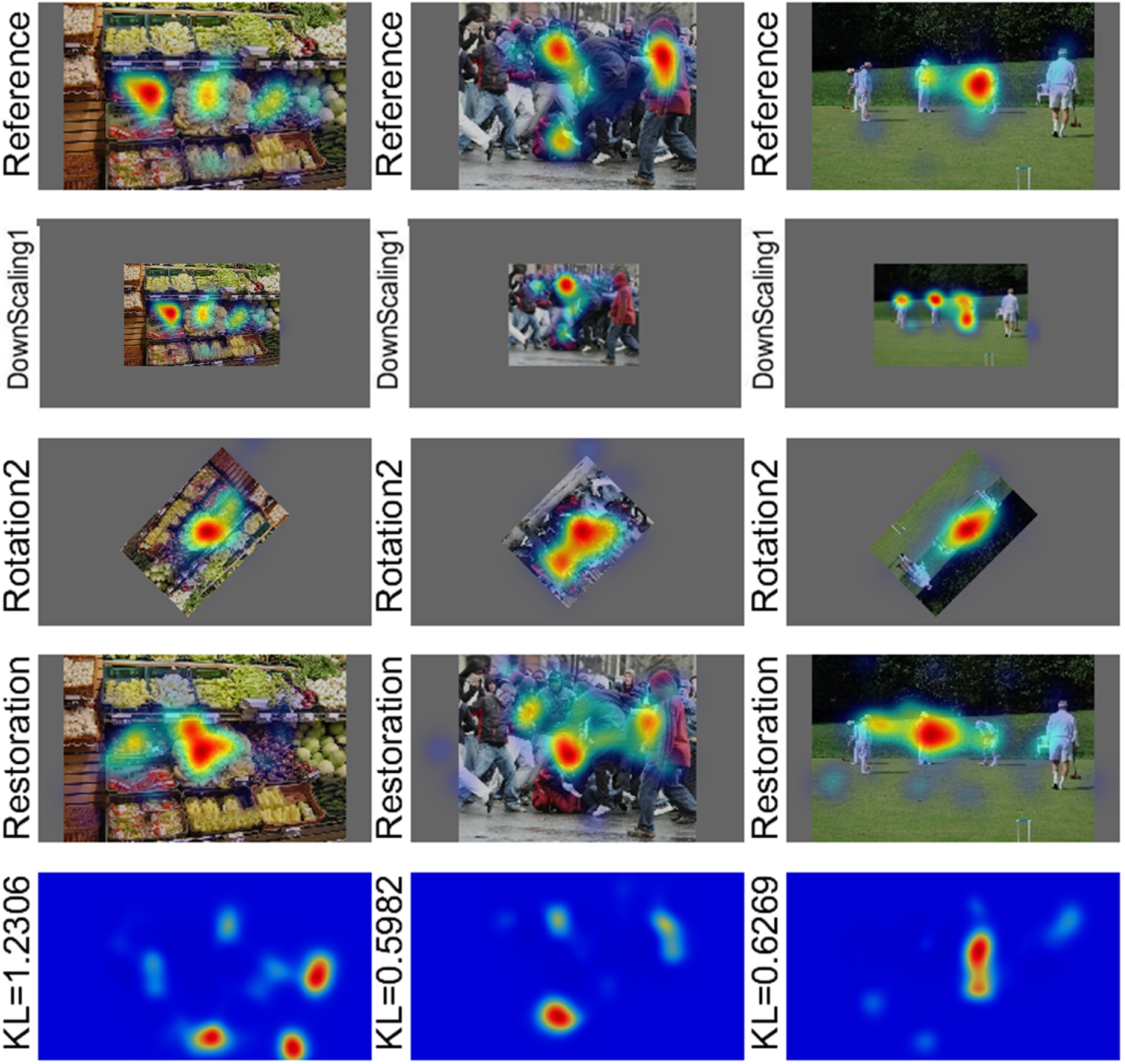}}
\vspace{-0.1cm}
\vspace{-0.1cm}
\caption{Human gaze discrepancies on Rotation2 compared to Reference. The $1_{st}$, $2_{nd}$ and $3_{rd}$ rows represent gaze maps of Reference, DownScaling1 and Rotation2, respectively. DownScaling1 serves as control groups here, because Rotation2 changes the effective size of the image compared to the Reference. The same scaling factor $\lambda_1=0.548$ is used for DownScaling1 and Rotation2 to mitigate the impact of image size on human gaze invariance. The $4_{th}$ row represents restored version of Rotation2 via inverse transformation. The restored version is aligned with Reference pixel-to-pixel for fair comparison.}
\label{Rotation2}
\end{figure}

\begin{figure}
\centering
\subfigure{\label{fig:edge-a}\includegraphics[height=0.8\linewidth]{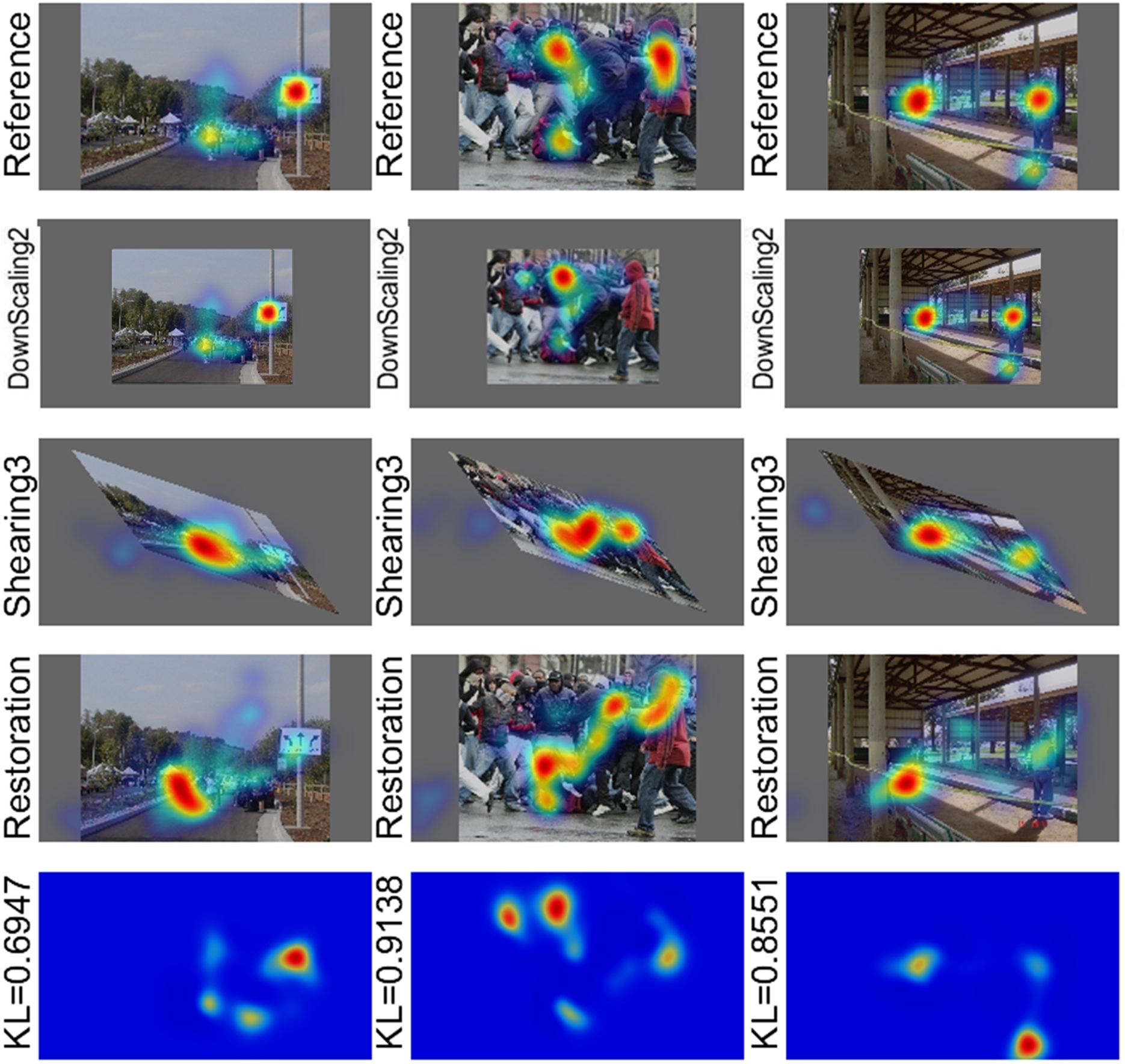}}
\vspace{-0.1cm}
\vspace{-0.1cm}
\caption{ Human gaze discrepancies on Shearing3 compared to Reference. The same scaling factor $\lambda_2=0.726$ is used for DownScaling2 and Shearing3 to mitigate the impact of image size on human gaze invariance.}
\label{Shearing3}
\end{figure}

\textbf{Rotation, Shearing:} Rotation and Shearing are spatial geometric transformations that alter original structural information and produce non-rigid objects.
As we can see in Fig.~\ref{Rotation2} and Fig.~\ref{Shearing3}, \textit{when viewing the rotated/affine-transformed stimuli, human gaze still focuses on semantic objects, but the intensities of the saliency regions are significantly changed by the geometric transformations.} For example, in the first column of Fig.~\ref{Shearing3}, when viewing Reference image, human gaze focuses on the ``guide board'' and ``pedestrians'', and the ``guide board'' attracts more human attention than ``pedestrians''. When viewing the affine-transformed image, although human fixations still locate at the ``guide board'' and ``pedestrians'' regions, the ``pedestrians'' attract more human attention. The similar cases can be observed in the $1_{st}$ and $3_{rd}$ columns of Fig.~\ref{Rotation2}, and the $3_{rd}$ column of Fig.~\ref{Shearing3}.

\textbf{Noise and Compression:} Noise and Compression are spatial perturbations that alter pixel intensities or texture, but maintain the structural information of the Reference image. Statistical comparison in Fig.~\ref{SMhuman} indicates that humans tolerate these spatial perturbations, demonstrating better invariance with regards to the Reference images.

\begin{figure}
\centering
\subfigure{\label{fig:edge-a}\includegraphics[height=0.46\linewidth]{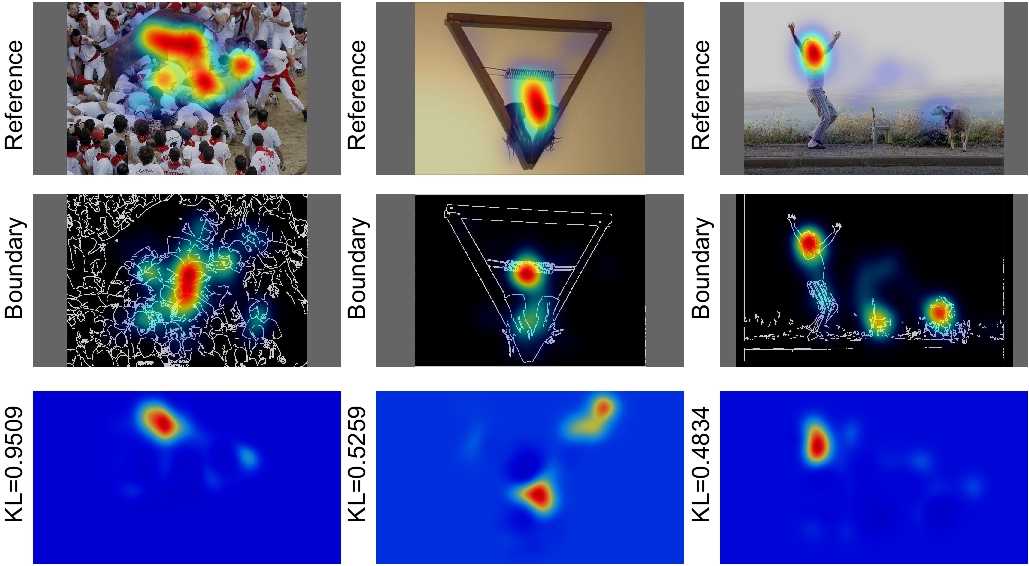}}
\vspace{-0.1cm}
\vspace{-0.3cm}
\caption{ Human gaze discrepancy on Boundary compared to the Reference.}
\label{Boundary}
\end{figure}

\begin{figure}
\centering
\subfigure{\label{fig:edge-a}\includegraphics[height=0.64\linewidth]{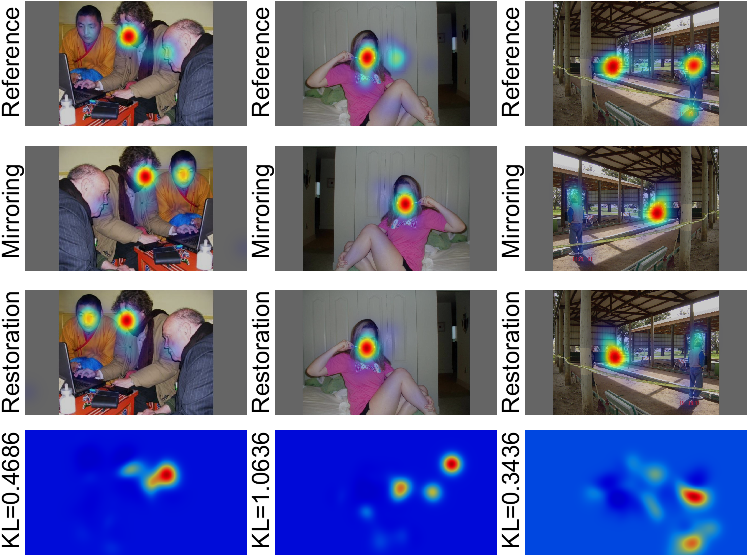}}
\vspace{-0.1cm}
\vspace{-0.1cm}
\caption{ Human gaze discrepancy on Mirroring compared to the Reference. The $3_{rd}$ row is the restored version of Mirroring via inverse transformation.}
\label{Mirroring}
\end{figure}

\textbf{Boundary:} Boundary transformation maintains most of the structural information of the Reference images, but lacks the texture, color and luminance information. As shown in Fig.~\ref{Boundary}, we notice that the semantic objects  still attract human gaze, \eg ``shoe'' and ``face'' in the $2_{nd}$ and $3_{rd}$ columns. However, for the scenes without clear semantic information, \eg the $1_{st}$ column, human gaze tends to focus on regions with sharp edges, thus causes discrepancy with the Reference image.
Statistical comparison in Fig.~\ref{SMhuman} indicates that Boundary transformation has sever impact on human gaze invariance compared to the spatial perturbations such as Noise and Compression, but results in better invariance than geometric transformations. Thus, we arrive at another empirical inference: \textit{For upright and rigid scenes, low-level structural and texture information helps to detect high-level salient regions}.

\textbf{Mirroring, Inversion:} Although Inversion is a special case of Rotation with $180\degree$ rotation angle, it demonstrates better invariance with Reference than geometric transformations. This is because Mirroring and Inversion are symmetric versions of Reference images and maintain both structural and texture information.
As shown in Fig.~\ref{Mirroring}, although human fixations on Mirroring and Reference have slight discrepancy on the trivial salient regions, they are consistent on major salient objects with obvious semantic information, such as ``face'' and ``pedestrians''.

Here, we list the lessons learned from our invariance analysis and the ways they can help saliency modeling as follows.
\begin{itemize}
\item \textbf{Discriminative semantic objects}: When a scene is cropped, human attention tends to focus on the salient regions with more semantic information that help to understand the cropped scene and to recover from the information loss.
\item \textbf{Highlighting semantic salient information while ignoring trivial artifacts}: We verified that human gaze focuses on semantic objects over various transformations, besides, human gaze tolerates the trivial artifacts caused by transformations such as JPEG and Noise distortions. In order to reach human level accuracy on transformed scenes, the robust saliency models should emphasize semantic salient regions while mitigating trivial artifacts.
\item \textbf{Leveraging structural and texture information}: For upright and rigid scenes, low-level structural and texture information helps to detect the salient regions.
\item \textbf{Combining multiple metrics}: There is no ``perfect'' metric that can accurately quantify human gaze on various transformations. However, they can complement each other. \footnote{Please see supplement for more details on the properties of different evaluation metrics on various transformations.}
\end{itemize}

Finally, we briefly discuss the impact of human attention invariance to other vision tasks such as object detection and classification. As we know that, region proposal has been successfully adopted in object detection \cite{fasterrcnn}. Saliency detection shares similar mechanism and goal with region proposal.
Besides, in classification task, top-down attention mechanism encodes semantic discriminative regions to boost classification convolution network \cite{attres}.
Different transformations will change the region proposal results at different levels. Wrong (or missing) region proposal will cause severe impact on final prediction of detection and classification applications.
Thus, the lessons via human attention analysis are generalizable to a plethora of attention-based detection and classification applications.
The robust approach should emphasize top-down semantic regions, and refine trivial bottom-up discriminative regions, in order to produce accurate region proposal.

\begin{figure}
\centering
\subfigure[\scriptsize sAUC$\uparrow$]{\label{fig:edge-a}\includegraphics[height=0.34\linewidth]{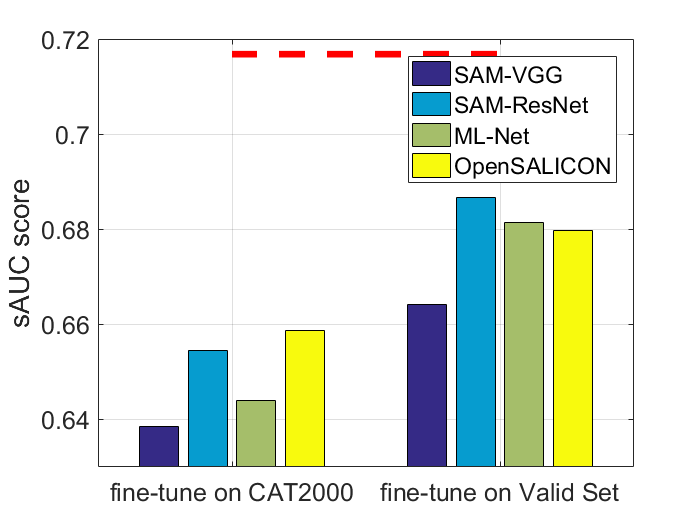}}
\subfigure[\scriptsize NSS$\uparrow$]{\label{fig:edge-a}\includegraphics[height=0.34\linewidth]{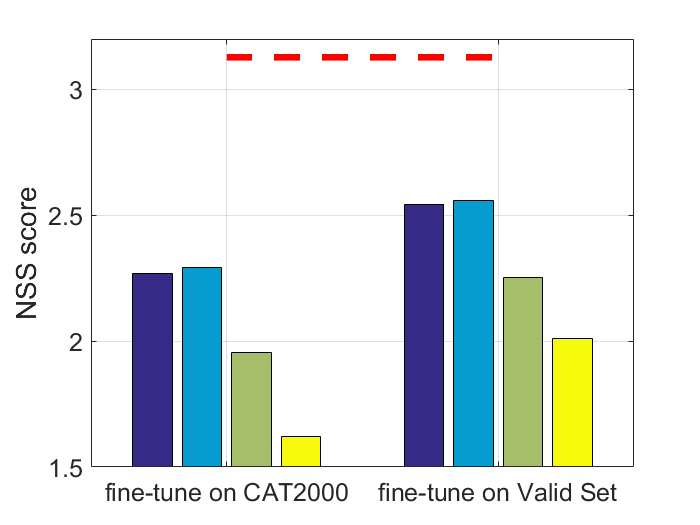}}\\ \vspace{-0.1cm}
\subfigure[\scriptsize sAUC$\uparrow$]{\label{fig:edge-a}\includegraphics[height=0.34\linewidth]{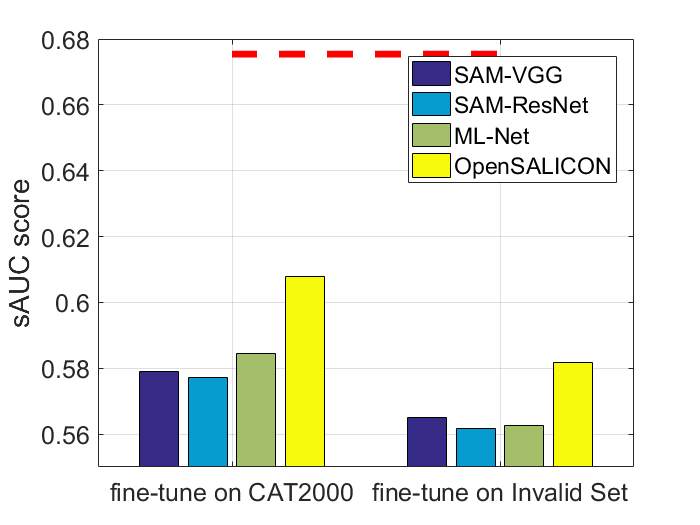}}
\subfigure[\scriptsize NSS$\uparrow$]{\label{fig:edge-a}\includegraphics[height=0.34\linewidth]{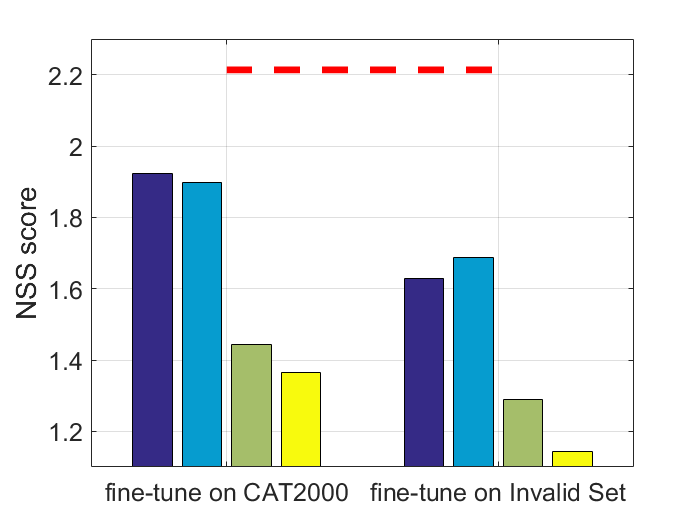}}\\ \vspace{-0.1cm}
\subfigure[\scriptsize sAUC$\uparrow$]{\label{fig:edge-a}\includegraphics[height=0.34\linewidth]{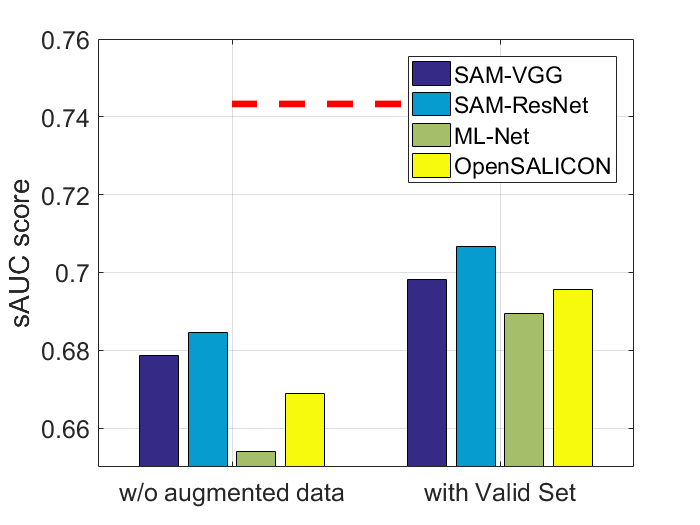}}
\subfigure[\scriptsize NSS$\uparrow$]{\label{fig:edge-a}\includegraphics[height=0.34\linewidth]{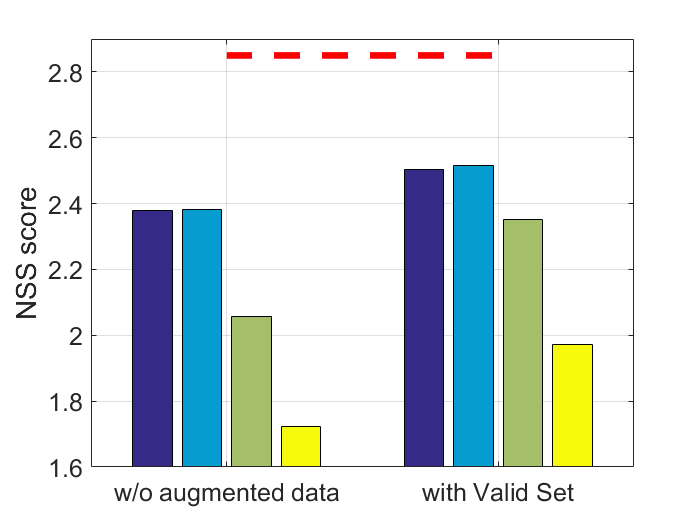}} \vspace{-0.1cm}
\caption{ Performances of 4 state-of-the-art deep saliency models on {\emph {valid}} ($1_{st}$ row) transformed set, {\emph {invalid}} ($2_{nd}$ row) transformed set, and distortion-free ($3_{rd}$ row) dataset. Notably, CAT2000 containing only distortion-free stimuli serves as a normal control group here.
The higher sAUC and NSS represent better performance.
The red dashed lines represent IO scores \cite{LSLGS} on each test set, which provide the upper-bound to prediction accuracy of objective models. We provide more results on CC and KL metrics in the supplement.}
\label{FineTunefinal}
\end{figure}

\section{Analysis of Data Augmentation}

The most common data augmentation strategy is to enlarge the training set using some label-preserving transformations, such as Cropping, Inversion, ContrastChange, and Shearing. However, different from classical image classification and object detection problems, the common data augmentation methods may produce label noise for the saliency prediction problem. This is because different transformations will change the ground truth at different levels. This work carries important implications as to which of these types of transformations are valid and which ones provide approximations of human gaze. We divide common transformations included in the proposed dataset into two sets: {\emph {valid}} and {\emph {invalid}} augmented sets, and explore how fine-tuning on different sets of augmented data can improve or degrade the performance of deep models with respect to ground truth.

On the one hand, we select Reference, Mirroring, Inversion, Contrast1, Shearing1, JPEG1 and Noise1 to generate a {\emph {valid}} augmented set, because these transformations have slight impacts on human gaze. On the other hand, Rotation1, Rotation2, Shearing2, Shearing3, Cropping1, Cropping2 and MotionBlur2 serve as an {\emph {invalid}} set, because these transformations are not able to preserve human gaze labels as approximations of the Reference.
We select 4 state-of-the-art deep saliency models, \ie SAM-VGG \cite{SAM}, SAM-ResNet \cite{SAM}, ML-Net \cite{MLnet}, and OpenSALCON \cite{SALICON}, for a comprehensive investigation.

We design and perform two experiments in this section:

\textbf{1}. Which types of transformations can improve the model robustness on distorted images?

\textbf{2}. Do the {\emph {valid}} augmentation transformations increase the model performance on normal distortion-free images?

\color{black}
\begin{figure*}[ht]
\centering
\includegraphics[scale=0.44]{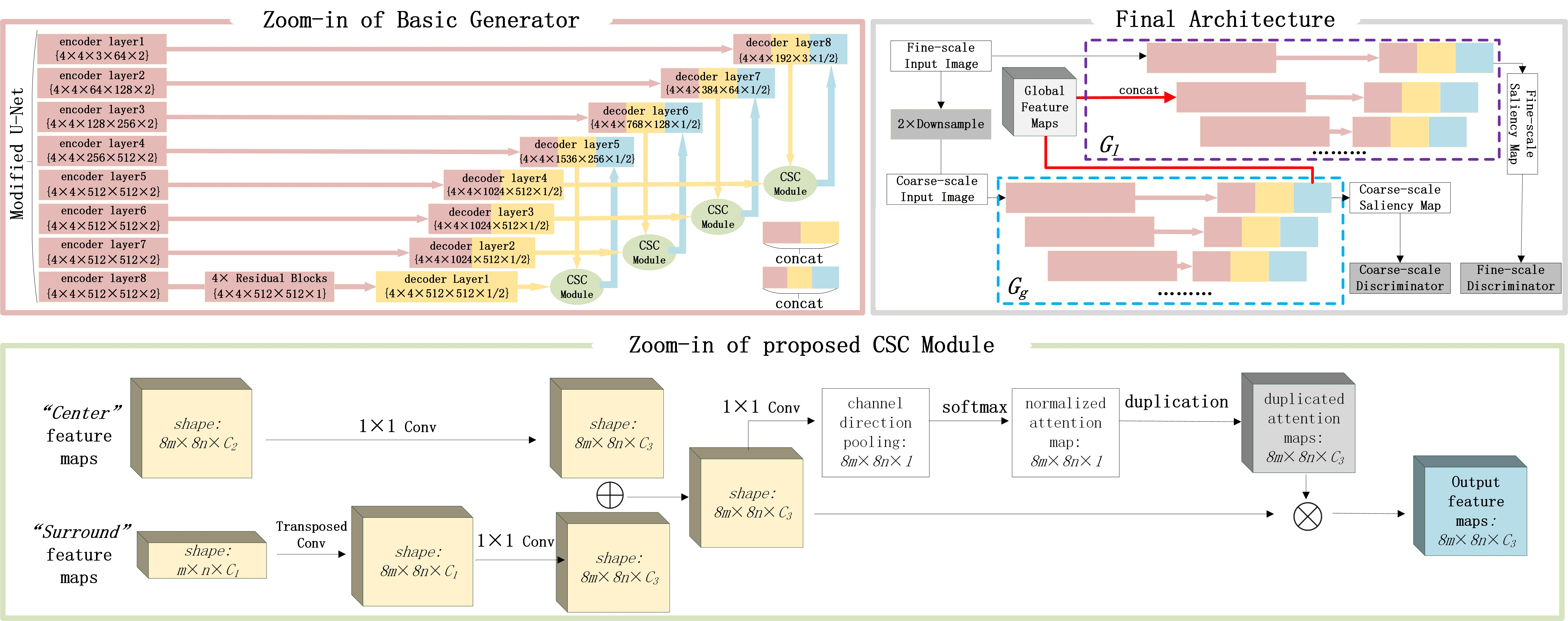} %scale=0.55
\caption{ Generator architecture of the proposed GazeGAN model. We represent the parameterization of convolution layer as \{height $\times$ width $\times$ input channel $\times$ output channel $\times$ stride\}. GazeGAN is equipped with a novel cross-scale short connection module, dubbed center-surround-connection (CSC). Proposed CSC module adopts a transposed convolution that learns to mitigate trivial artifacts in upsampling stage. Besides, CSC module also utilizes the element-wise summation and attention mechanism to emphasize semantic information. Proposed CSC module is generalizable to any encoder-decoder CNN architecture.} %
\label{Architecture}
\vspace{-0.4cm}
\end{figure*}

In the first experiment, we select some distortion-free images from the CAT2000 dataset as a normal control group, because the proposed dataset has similar content with CAT2000, such as indoor, outdoor, fractals and cartoon images.
Specifically, each of {\emph {valid}}, {\emph {invalid}} and normal control group is divided into a training set (550 images) and a test set (150 images), respectively. We borrow 100 images from CAT2000 as validation set for selecting optimal hyper-parameters.

In the first experiment, the model training process includes two steps, \ie pre-training and fine-tuning. First, each model is pre-trained on SALICON dataset. This dataset contains 10,000 training images, 5,000 validation images and 5,000 test images. Next, we fine-tune the pre-trained models on 3 different datasets, \ie \textit{valid} transformed set, \textit{invalid} transformed set, and distortion-free CAT2000 set, as shown in the $1_{st}$ and $2_{nd}$ rows of Fig.~\ref{FineTunefinal}.

In the second experiment, we select 1500 distortion-free images from CAT2000 as original training set, 400 images as test set, and 100 images as validation set. Then, we use the {\emph {valid}} transformations to enlarge the original training set of CAT2000 to 10500 images. Similarly, the deep models are first pre-trained on SALICON training set. We then fine-tune the pre-trained models using the augmented CAT2000 training set (10500 images) and the original CAT2000 training set w/o augmented data (1500 images), respectively. Performance comparisons of these two fine-tuning strategies are shown in the $3_{rd}$ row of Fig.~\ref{FineTunefinal}.

For fair comparison, we unify the experimental setup for different data augmentation strategies.
In the pre-training stage, we set the training hyper-parameters as follows: 1) For the 4 deep models mentioned in Fig.~\ref{FineTunefinal}, stochastic gradient descent (SGD) serves as the optimization function with momentum of 0.9, weight decay of 0.0005, and the batch size of 1, and 20 training epochs, 2) For the ML-Net, learning rate is $\rm 10^{-2}$, 3) For OpenSALICON, learning rate is $\rm 10^{-6}$, and 4) For SAM-VGG and SAM-ResNet, initial learning rates are set to $\rm 3\times10^{-5}$, and are decreased by 10 every two epochs for SAM-ResNet, and every three epochs for SAM-VGG.
In the fine-tuning stage: 1) We also adopt SGD with momentum of 0.9 and weight decay of 0.0005, and set batch size to 1, fine-tuning epoch to 10, 2) For ML-Net, learning rate is $\rm 10^{-3}$, 3) For OpenSALICON, learning rate is $\rm 10^{-7}$, and 4) For SAM-VGG and SAM-ResNet, initial learning rates are $\rm 3\times10^{-7}$, and are decreased by 10 every two epochs for SAM-ResNet, and every three epochs for SAM-VGG.

Experimental results shown in the $1_{st}$ row of Fig.~\ref{FineTunefinal} verify that fine-tuning using the {\emph {valid}} transformed set can improve deep models' robustness on the distorted test set, compared to using CAT2000 which contains only distortion-free images. However, as shown in $2_{nd}$ row of Fig.~\ref{FineTunefinal}, fine-tuning using the {\emph {invalid}} transformed set degrades deep models' performances compared to using normal stimuli. The results of the $3_{rd}$ row of Fig.~\ref{FineTunefinal} indicate that the {\emph {valid}} transformations provide an efficient data-augmentation approach to utilize expensive eye-movement data for boosting deep saliency models.

\section{The Proposed GazeGAN Model}

We recall the lessons learned from human gaze analyses (\ie Section-III), and list the general ideas behind the proposed model as follows:
\vspace{-5pt}
\begin{itemize}
    \item {\bf Conditional GAN (for discriminating semantic object):} The generator aims to fool the discriminator that is trained to distinguish synthetic saliency maps from real human gaze. The discriminator conditioned by the transformed images can boost generator to focus on semantic salient objects as real human gaze;
    \item {\bf Center-surround connection (for highlighting semantic information, while mitigating trivial artifacts):} Inspired by human visual center-surround antagonism mechanism, we propose a novel cross-scale short connection module, which helps model output to mitigate wrong predictions caused by trivial artifacts, while concentrating on semantic salient objects, in order to reach human level accuracy on transformed scenes; % [[rephrase! unclear. ]]
    \item {\bf Skip-connections (for leveraging structural and texture information):} Skip-connections  combine low-level structural and texture features from encoder layers with high-level semantic features from decoder layers, because the low-level features also help to detect salient regions;
    \item {\bf Local-global GANs (more robust to scale transformation):} Multiple generators learn different groups of spatial representations in different scales, while multiple discriminators can improve the intermediate prediction results from coarse to fine.
\end{itemize}

\subsection{The generator}

As shown in Fig.~\ref{Architecture}, the backbone GazeGAN generator is a modified U-Net equipped with a novel ``center-surround connection module'' (CSC module).

U-Net is a powerful fully convolutional network presented by Olaf~\etal \cite{UNET}. It has made a great breakthrough in biomedical image segmentation by predicting each pixel's class. In saliency prediction, the goal of U-Net is predicting each pixel's probability of being salient. Compared to the
generator of SalGAN \cite{SalGAN} saliency model (\ie VGG-16), U-Net consists of symmetric encoder and decoder layers, and utilizes skip connections to combine low-level structural and texture features from encoder layers with high-level semantic features from decoder layers.

An important early vision mechanism in the human vision system that serves recognition and attention is the ``center-surround'' mechanism. The early visual neurons (retina and LGN) are most sensitive in a small region of the visual space (\ie $\emph{center}$ of receptive field), while stimuli presented in the antagonistic region concentric to the $\emph{center}$ (the $\emph{surround}$) inhibit the neuronal response \cite{IttiKoch}. The ``Center-surround'' mechanism highlights local spatial discontinuities and is well-suited for detecting salient locations that stand out from their $\emph{surround}$ while suppressing other trivial information, such as noise and artifacts.

For improving the robustness of deep saliency models, we add the ``center-surround'' mechanism into the CNN model for the first time.
Here, we implement the ``center-surround'' operation as a cross-scale short connection module, because it is generalizable to any encoder-decoder CNN architecture, as shown in Fig. \ref{Architecture}.
Specifically, we select the feature maps in a coarse scale (the $\emph{surround}$) from the $i_{th}$ decoder layer, where $i \in \{1,2,3,4\}$, and the corresponding fine scale maps (the $\emph{center}$) are from the $j_{th}$ decoder layer, where $j \in \{i + 4 \}$. We first use a $3 \times 3$ transposed convolution layer to upsample the $\emph{surround}$ feature maps to have the same resolution (height $\times$ width) with the $\emph{center}$ maps. Besides, in the upsampling stage, this transposed convolution also learns to reduce the wrong predictions caused by trivial artifacts. Next, we employ the 1$\times$1 convolution layers to unify the channels of \emph{center} and \emph{surround} maps while keeping the resolution fixed.
Then, we compute the preliminary $\emph{center-surround}$ output by an element-wise summation as:
\vspace{-0.1cm}
\begin{equation}
\small
f_{cs}^{i,j} = (\mathcal N \ast (\mathcal U \ast f_s^i)) \oplus (\mathcal N \ast f_c^j),
\label{nonliearsub}
\end{equation}
where $f_s^i$ and $f_c^j$ represent the \emph{surround} feature maps of the $i_{th}$ layer, and the \emph{center} feature maps of the $j_{th}$ layer, respectively.
$\mathcal N$ represents the $1\times1$ convolution,
and $\mathcal U$ represents the transposed convolution.
$f_{cs}^{i,j}$ represents the preliminary \emph{center-surround} response of $f_s^i$ and $f_c^j$. $\oplus$ is an element-wise summation. $\ast$ is the convolution operation.

Next, we utilize the attention mechanism to further highlight the semantic saliency regions detected by $f_{cs}^{i,j}$. Specifically, we feed the $f_{cs}^{i,j}$ into a $1\times1$ convolution to squeeze the channel amount as 1, thus obtain a 2D one-channel map $\bar f_{cs}^{i,j}$. Then, we compute the 2D normalized attention map of $\bar f_{cs}^{i,j}$ via softmax function:
\begin{equation}
\small
\left\{
             \begin{array}{lr}
             \bar f_{cs}^{i,j} = \mathcal N \ast f_{cs}^{i,j},\\
             A_{cs}^{i,j} = \textup{softmax}(\bar f_{cs}^{i,j}) = \frac{\textup{exp}(\bar f_{cs}^{i,j}(m,n))}{\sum_m\sum_n \textup{exp}(\bar f_{cs}^{i,j}(m,n))},\\
             \end{array}
\right.
\label{atten}
\end{equation}

The final output of the CSC module is computed by the element-wise product of $f_{cs}^{i,j}$ and normalized attention map as $\widetilde{f_{cs}^{i,j}} = f_{cs}^{i,j} \otimes \widetilde{A_{cs}^{i,j}}$, where $\widetilde{A_{cs}^{i,j}}$ is the expanded 3D attention map via duplicating the 2D map $A_{cs}^{i,j}$ in channel direction.

Finally, we concatenate the obtained \emph{center-surround} maps $\widetilde{f_{cs}^{i,j}}$ with the other feature maps from the $j_{th}$ decoder layer in channel direction. This way, each of the $5_{th}-8_{th}$ decoding layers concatenate 3 types of feature maps, \ie $[f_{st}^{k}, f_{sm}^{j}, \widetilde{f_{cs}^{i,j}}]$. Specifically, $f_{st}^{k}$ is low-level structural/texture features from the $k_{th}$ encoder layer via skip-connection, $f_{sm}^{j}$ is semantic features from the $j_{th}$ decoder layer, and $\widetilde{f_{cs}^{i,j}}$ is \emph{center}-\emph{surround} features via CSC modules that mitigate trivial artifacts. %thus demonstrating better robustness against transformations.
In Fig.~\ref{Architecture}, we use red, yellow, and blue rectangles to represent $f_{st}^{k}, f_{sm}^{j}, \widetilde{f_{cs}^{i,j}}$, respectively. Notice that all activation functions in encoder layers are leaky-ReLUs with slope $=$ 0.2, while activation functions in decoder layers are normal ReLUs.

In the final architecture shown in Fig.~\ref{Architecture}, we further append a local generator $G_l$ on basis of the global generator $G_g$, in order to extract more high-resolution features. $G_g$ is able to detect the fine-scale semantic salient objects, while $G_l$  encodes more salient objects in coarse scales, \ie small face and tiny text.
Specifically, we concatenate the feature maps from the last decoder layer of $G_g$ with the feature maps from the second encoder layer of $G_l$, to integrate the global semantic information from coarse to fine. We feed the original image into the $G_l$, and feed the downsampled image into the $G_g$. The $G_g$ and $G_l$ are jointly trained end to end.

\begin{figure}
\centering
\subfigure[\scriptsize Conditional discriminator]{\label{fig:edge-a}\includegraphics[height=0.5\linewidth]{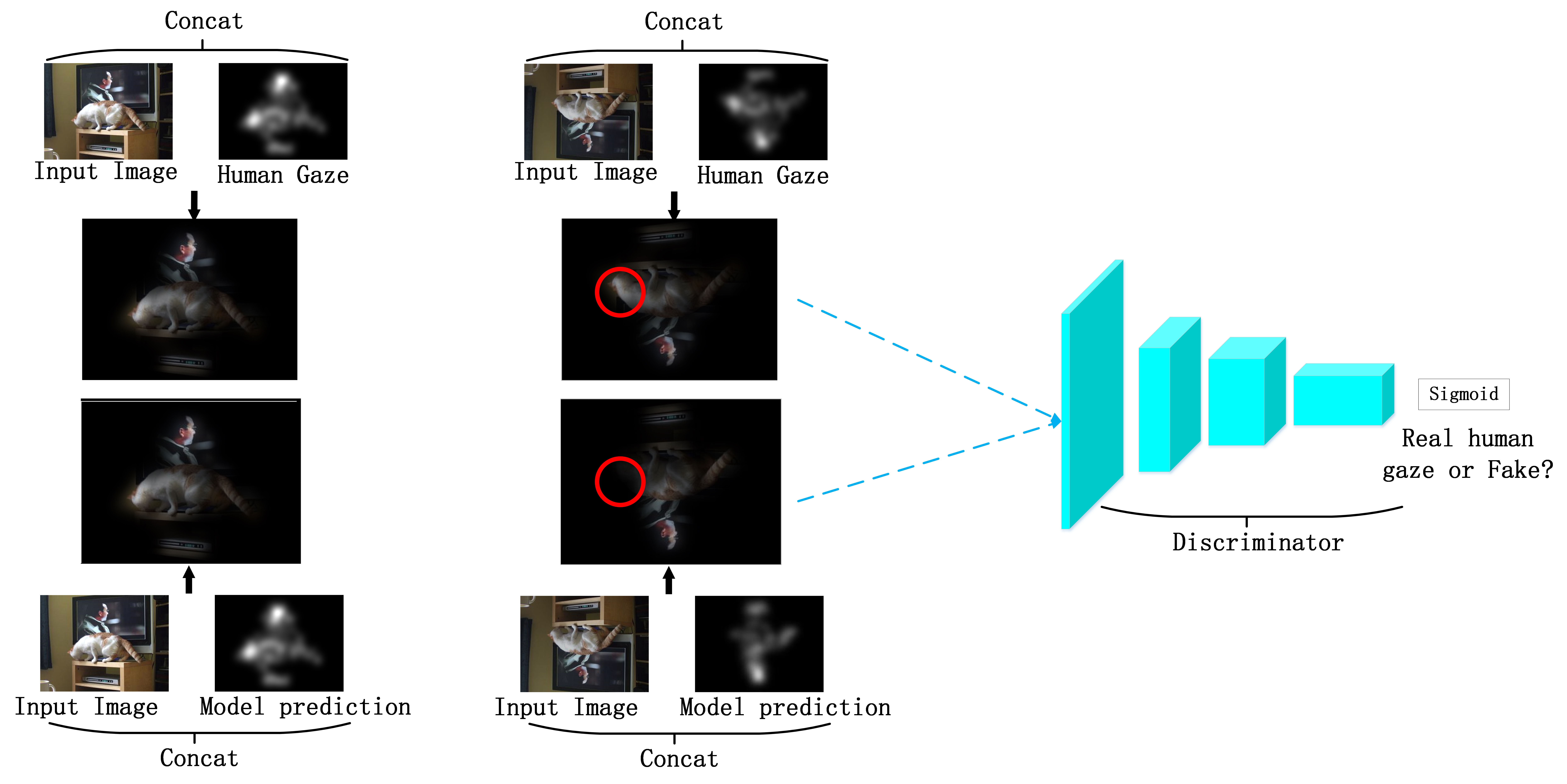}}
\subfigure[\scriptsize Normal discriminator]{\label{fig:edge-a}\includegraphics[height=0.35\linewidth]{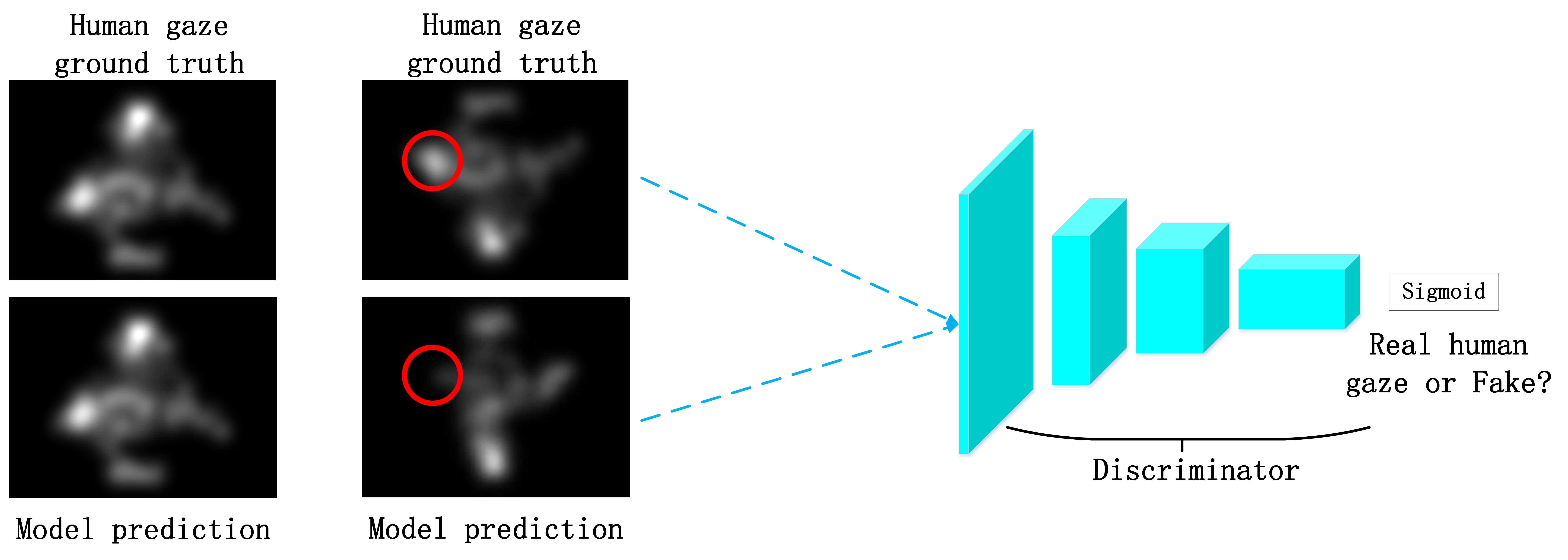}}
\vspace{-0.1cm}
\vspace{-0.1cm}
\caption{ (a) Conditional discriminator whose inputs are ``Image $\&$ Saliency Map'' pairs, which has access to the (transformed) input images, demonstrating better discriminating ability on semantic object. (b) Normal discriminator whose inputs are only saliency maps.}
\label{ArchitectureDis}
\end{figure}

\subsection{The discriminator}

To discriminate real human gaze from synthetic saliency map, we train a 5-layer patch-based discriminator \cite{imgtoimg}, which contains 4 convolution layers with increasing number of $4 \times 4$ convolution kernels, increasing by a factor of 2 from 64 to 512 kernels. On top of the 512 feature maps generated by the discriminator layer4, we append a sigmoid layer with $4 \times 4$ filter kernels and sigmoid activation function to obtain the final probability of being the real human gaze. Notice that we concatenate the saliency map (or human gaze) with original input color image in channel direction, and feed them to the discriminator simultaneously. Thus, GazeGAN is a conditional GAN \cite{imgtoimg} because both the generator and the discriminator can observe the input source image, as shown in Fig.~\ref{ArchitectureDis}. Particularly, the conditional discriminator has access to both input images (including transformation type) and the corresponding saliency maps, demonstrating better discrimination ability on semantic objects than the normal discriminator.
We append the conditional discriminators to the end of $G_g$ and $G_l$, respectively, in order to improve the predictions from coarse to fine.

\subsection{Loss functions}

In the human gaze analysis section, we found that there is no ``perfect'' evaluation metric that can accurately quantify human gaze on various transformations. However, metrics can compensate for each other.}
Previous works \cite{imgtoimg,srgan,deblurgan} have proved it beneficial to mix the \textit{adversarial loss} with some task-specific \textit{content losses} to train a GAN.

\subsubsection{The content loss}

For saliency detection task, it has been proved that a linear combination of different saliency evaluation metrics achieves a good performance \cite{SALICON,SAM}.

CC, KL and NSS \cite{nssdef} metrics \footnote{See supplement for more details about KL, CC, and NSS losses.} perform well in measuring the pixel-level similarity between ground-truth and synthetic maps. However, we found that only using a linear combination of pixel-level losses produce high discrepancy between the grey-level histograms of synthetic result and human gaze. \footnote{Please see supplement for visualization of this issue.}

For solving the drawbacks of pixel-level losses, we propose a histogram loss to reduce the histogram discrepancy between the generated saliency map and the human gaze map. The histogram loss includes two steps, {\emph{i.e.}} histogram distribution estimation and histogram similarity calculation.
For constructing a differentiable histogram loss, we first devise the histogram estimation method based on Ustinova's work \cite{nipshistloss}.
We denote the pixel luminance of saliency map as $l_i$, $i \in [1, S]$, where $S$ represents the number of pixels in the saliency map.
Suppose that the distribution of $l_i$ is estimated as the $(N+1)$-dimensional histogram with the nodes $b_0$ = 0, $b_1$ = $\frac{255}{N} \times 1$, ..., $b_{N}$ = 255 uniformly filling $[0, 255]$ with the step $\Delta = \frac{255}{N}$. Then, we use equation \ref{HistPi} to estimate the probability distribution (denoted as $p_k$, where $k \in [0, N]$) for each node of the histogram.

\begin{equation}
\small
p_k=\frac{1}{S}\times(\sum_{l_i \in [b_{k-1}, b_k)}\frac{l_i - b_{k-1}}{\Delta} + \sum_{l_i \in [b_{k}, b_{k+1}]}\frac{b_{k+1} - l_i}{\Delta}),
\label{HistPi}
\end{equation}

We then adopt the {\textit{min-max}} normalization method to normalize $p_k$ as $\bar{p_k}$, to guarantee that $\bar{p_k} \in [0,1]$.
Next, we utilize the Alternative Chi-Square (ACS) distance to measure the histogram similarity.\footnote{Derivative of proposed $L_{ACS}$ loss is provided in Supplementary Material}

\begin{equation}
\small
L_{ACS} = 2 \times \sum_{k=0}^{N} \frac{(\bar{p_k} - \bar{q_k})^2}{\bar{p_k} + \bar{q_k} + \epsilon},  %\frac{\bar{p_k^{SM}}{1}
\label{histloss}
\end{equation}
where $\bar{p_k}$ and $\bar{q_k}$ represent the normalized probability distribution at the $k_{th}$ node of histograms of generated saliency map and ground-truth human gaze, respectively. $\epsilon = 10^{-8}$ is a smoothing term to avoid division by zero. We set $N$ to 255.

As shown in equation \ref{finalcontent}, the final content loss $L_{cont}$ is a linear combination of four pixel-level losses $\rm L_1$, KL, CC and NSS, and a histogram loss $L_{ACS}$. In Section-$\rm \uppercase\expandafter{\romannumeral6}$, we quantify the contribution of each loss function via ablation study.
\begin{equation}
\small
\begin{split}
&  L_{cont} =  w_1 \textup L_1( GT_{den},  SM) +  w_2 \textup{KL}(GT_{den}, SM) + \\
& w_3 \textup{CC}(GT_{den}, SM) + w_4 \textup{NSS}( GT_{fix},  SM) + w_5 L_{ACS}.%(GT_{den}, SM)
\end{split}
\label{finalcontent}
\end{equation}
where $w_i$, $i \in \{1,2,3,4,5\}$ are five scalars to balance five losses, and the good default settings are 1, 10, -2, -2 and 1, respectively. The good default scalars are tested and selected via SALICON validation set. The smaller values for $\rm L_1$, KL and $L_{ACS}$ scores indicate higher similarity between synthetic result and ground-truth, whereas for CC and NSS, the higher values indicate higher similarity.

\subsubsection{The adversarial loss}
The adversarial loss $L_{adv}$ is expressed as
\begin{equation}
\small
\begin{split}
%L_{adv}(G, D) = \mathbb{E}_{I, GT_{den}}[\log D(I, GT_{den})] +
& L_{adv}(\textup{\bf G}, \textup{\bf D}) = \mathbb{E}_{I, GT_{den}}[\log \textup{\bf D}(I, GT_{den})] + \\
& \mathbb{E}_{I, \textup{\bf G}(I)}[\log(1 - \textup{\bf D}(I, \textup{\bf G}(I)))], \\
\end{split}
\label{advloss}
\end{equation}
where $I$ means the original input image, while $\textup{\bf G}$ and $\textup{\bf D}$ represent generator and discriminator. $\textup{\bf G}$ represents the global and local generators (\ie $G_g$ and $G_l$), while $\textup{\bf D}$ represents the fine-scale and coarse-scale discriminators.
$\textup{\bf G}$ tries to minimize this adversarial loss against an adversarial $\textup{\bf D}$ which tries to maximize it, {\emph{i.e.}} $\rm arg$ $\rm min_{\textup{\bf G}}max_{\textup{\bf D}}$ $L_{adv}(\textup{\bf G}, \textup{\bf D})$. %$\rm arg$ $\rm min_{\textsl \textup{G}}max_{\textit D}$ $L_{adv}(G, D)$

\section{Experiments and Results}
%In this section, we introduce the experimental settings, and perform an ablation analysis to validate the contribution of each component of our model. Finally we compare the proposed model with state-of-the-art.
\subsection{Experimental setup}
We use 4 datasets to ensure a comprehensive comparison including: \textbf{1)} SALICON dataset (previously released) \cite{SALICONDB}; \textbf{2)} LSUN'17 dataset (SALICON-2017-released-version) \cite{LSUN17}; \textbf{3)}  MIT1003 dataset \cite{juddiccv} and \textbf{4)} The proposed dataset.

For SALICON, MIT1003, and LSUN'17 datasets, we resize input images to $\rm 480\times640$ for saving computing resources. Considering that the images of MIT1003 have different resolutions, we apply zero padding bringing images to have a unified aspect ratio of 4:3 and resize them to have the same size. Images of the proposed dataset have the same input size of $\rm 1080\times1920$, hence we resize them to $\rm 360\times640$.

For fair comparison, all of the deep-learning based models are trained from scratch on the SALICON (previously released) dataset. Specifically,
we first adopt the proposed \textit{valid} data augmentation transformations to enlarge the 10,000 training images. This way, we obtain another 60,000 augmented stimulus set with 6 types of label-preserving transformations.
For SALICON \cite{SALICON}, SAM-VGG \cite{SAM}, SAM-ResNet \cite{SAM} and SalGAN \cite{SalGAN} models, we follow their authors' guideline to initialize their network parameters using the pre-trained weights on ImageNet \cite{imagenetdataset}.  The proposed GazeGAN is initialized from a Gaussian distribution with mean 0 and standard deviation 0.02,
which achieves similar performance with the ImageNet initialization method.
We use the augmented 70,000 training samples to train all of the competing models. We select 4,000 images
from the SALICON validation set as the test set and the remaining 1,000 images serve as the validation set for selecting the optimal hyper-parameters.

For MIT1003 dataset, we randomly divided it into a training set with 600 images, a validation set with 100 images, and a test set with 303 images. We use the same data augmentation method to enlarge the training set of MIT1003 dataset. For all competing models, we reload the parameters pre-trained on the augmented SALICON training set. We then fine-tune the models on the augmented MIT1003 training set.

The proposed dataset consists of 19 transformation groups, and each group contains 100 images. We divide each group into 60 training images, 10 validation images and 30 test images. This way, we obtain 1140 training samples, 190 validation samples and 570 test samples. Similarly, for all competing models, we reload the parameters pre-trained on the augmented SALICON training set, then we fine-tune the models on 1140 training samples of proposed dataset.

For LSUN'17 dataset, the performance scores of other competing models are from LSUN'17 SALICON Saliency Prediction Competition system \cite{LSUN17}, where our model is under the username ``\textit{codacscgaze}''.

In the training stage,
we encourage the generator of the proposed GazeGAN to minimize the linear combination of the content loss $L_{cont}$ and the adversarial loss $L_{adv}$. Besides, rather than training the discriminator to maximize $L_{adv}$, we instead minimize -$L_{adv}$.
Adam optimizer \cite{adamopt} with a fixed learning rate $\rm lr = 2\times10^{-4}$, and the momentum parameter of $\beta_1 = 0.5$ serves as the optimization method to update the model parameters. We alternatively update the generators and discriminators as suggested by Goodfellow~\etal \cite{goodfollowGAN}. The batch-size is set as 1.
Our implementation is based on Pytorch and Tensorflow flowcharts, using NVIDIA Tesla GPU.

\begin{table}
\centering
\renewcommand{\arraystretch}{1.5}
\renewcommand{\tabcolsep}{.3mm}
\small
    \caption{Ablation study addressing using different loss functions on LSUN'17 (SALICON-2017-Version) validation set.}
    \label{tab:AblaLoss}
    \scriptsize
    \centering
    \begin{tabular}{|c|c|c|c|c|c|c|c|}

    \hline
    {\bfseries Dataset} &{\bfseries Loss functions} & \bfseries CC$\uparrow$ & \bfseries NSS$\uparrow$ & \bfseries sAUC$\uparrow$  & \bfseries KL$\downarrow$ \\
   \hline
   \hline

    \multirow{2}*{LSUN'17}
        ~ &{$\rm L_1$ + KL + CC + NSS} & {0.823} &{  1.388} &{  0.686}  &{  0.876} \\
    ~ &{$\rm L_1$ + KL + CC + NSS + $L_{adv}$} & {0.849} &{  1.493} &{  0.718}  &{  0.587} \\
    \cline{2-2}
    ~ &{$\rm L_1$ + KL + CC + NSS + HistLoss} & {0.855} &{  1.557} &{  0.712}  &{  0.606} \\
    ~ &{$\rm L_1$ + KL + CC + NSS + HistLoss + $L_{adv}$}  &{ \bfseries 0.881} &{  \bfseries 1.911} &{  \bfseries 0.738}  &{  \bfseries 0.373} \\

   \hline
    \end{tabular}
%\vspace{-5pt}
\end{table}

\begin{table}
\centering
\renewcommand{\arraystretch}{1.5}
\renewcommand{\tabcolsep}{1.0mm}
    \caption{Ablation study of different modules of GazeGAN on LSUN'17 (SALICON-2017-Version) validation set. $V_1$-$V_4$ are four different variations made up of different modules.}
    \label{tab:AblaModule}
    \scriptsize
    \centering
    \begin{tabular}{|c|c|c|c|c|c|c|c|}
    \hline
    {\bfseries Dataset} &{\bfseries Component Modules} &{\bfseries CC$\uparrow$} & {\bf NSS$\uparrow$} & {\bf sAUC$\uparrow$}  & {\bf KL$\downarrow$} \\
   \hline
   \hline
        \multirow{2}*{LSUN'17} &{$V_1$: Plain U-Net} & {0.752} &{ 1.221} &{ 0.613}  &{ 0.824} \\
    ~ &{$V_2$: $V_1$ + Residual blocks} &{ 0.849} &{ 1.472} &{ 0.689}  &{ 0.530} \\
    ~ &{$V_3$: $V_2$ + CSC Module} & {0.865} & {1.609} & {0.718}  & {0.489} \\
    ~ &{$V_4$: $V_3$ + Local Generator} &{ \bfseries 0.881} &{  \bfseries 1.911} &{  \bfseries 0.738}  &{  \bfseries 0.373} \\

    \hline
    \end{tabular}
\end{table}

\begin{table}
\centering
\renewcommand{\arraystretch}{1.5}
\renewcommand{\tabcolsep}{1.5mm}
    \caption{Performance comparison on test set of LSUN'17 competition (SALICON-2017-version).}
    \label{Test2017}
    \scriptsize
    \centering
    \begin{tabular}{|c|ccccccc|}

    \hline
    {\bfseries } &{\bfseries sAUC$\uparrow$} &{\bfseries IG$\uparrow$} &{\bfseries NSS$\uparrow$} &{\bfseries CC$\uparrow$} &{\bfseries AUC$\uparrow$} &{\bfseries SIM$\uparrow$} &{\bfseries KL$\downarrow$} \\
   \hline
   \hline

    \hline
    {{ \bfseries GazeGAN}} &{ 0.736} &{\bfseries 0.720} &{ 1.899} &{ 0.879} &{ 0.864} &{ 0.773} &{ \bfseries 0.376}\\
    \hline

     \hline
    {{ SAM-ResNet}} &{0.741} &{0.538} &{1.990} &{0.899} &{0.865} &{0.793} &{0.610}\\
    \hline

    \hline
    {{ EML-Net}} &{\bfseries 0.746} &{0.716} &{\bfseries 2.050} &{0.886} &{\bfseries 0.866} &{0.780} &{0.520}\\
    \hline

    \hline
    {{ DI-Net}} &{0.739} &{0.195} &{1.959} &{\bfseries 0.902} &{0.862} &{\bfseries 0.795} &{0.864}\\
    \hline

     \hline
    {{ CEDNS}} &{0.745} &{0.357} &{2.045} &{0.862} &{0.862} &{0.753} &{1.026}\\
    \hline

    \hline
    {{ lvjincheng}} &{0.726} &{0.613} &{1.829} &{0.856} &{0.855} &{0.705} &{\bfseries 0.376}\\
    \hline

    \hline
    {{ hallazie}} &{0.724} &{0.640} &{1.804} &{0.844} &{0.855} &{0.714} &{0.381}\\
    \hline

    \hline
    {{ RyanLui}} &{0.724} &{-0.187} &{1.838} &{0.855} &{0.850} &{0.746} &{1.208}\\
    \hline

    \hline
    {{ hrtavakoli}} &{0.717} &{0.541} &{1.773} &{0.848} &{0.845} &{0.684} &{0.492}\\
    \hline

    \hline
    {{ sfdodge}} &{0.720} &{0.646} &{1.911} &{0.821} &{0.856} &{0.722} &{0.527}\\
    \hline

     \hline
    \end{tabular}
\end{table}

\begin{table}
\centering
\renewcommand{\arraystretch}{1.5}
\renewcommand{\tabcolsep}{1.3mm}
\scriptsize
    \caption{Performance on MIT1003 dataset \cite{juddiccv}. For fair comparison, all competitors are fine-tuned on MIT1003 training set.}
    %\vspace{-10pt}
    \label{tab:perform2}
    \scriptsize
    \centering
    \begin{tabular}{|c|cccccc|}

    \hline
    { } &{\bfseries AUC-Judd$\uparrow$} &{\bfseries CC$\uparrow$} &{\bfseries NSS$\uparrow$}  &{\bfseries sAUC$\uparrow$} &{\bfseries SIM$\uparrow$} &{\bfseries KL$\downarrow$} \\
   \hline
   \hline

    \hline
    {{ SAM-ResNet \cite{SAM}}} &{ 0.880} &{0.649} &{\bfseries 2.439}  &{0.748} &{0.447} &{1.092}\\
    \hline

    \hline
    {{ \bfseries GazeGAN}} &{\bfseries 0.883} &{\bfseries 0.654} &{2.402}  &{0.747} &{ 0.446} &{\bfseries 1.042}\\
    \hline

    \hline
    {{ DVA \cite{dvatip}}} &{0.870} &{0.640} &{2.380}  &{\bfseries 0.770} &{\bfseries 0.500} &{1.120}\\
    \hline

    \hline
    {{ SAM-VGG \cite{SAM}}} &{0.880} &{0.643} &{2.377}  &{0.740} &{0.415} &{1.141}\\
    \hline

   \hline
    {{ OpenSALICON \cite{SALICON}}} &{0.864} &{0.639} &{2.140}  &{0.742} &{0.434} &{1.136}\\
    \hline

     \hline
    \end{tabular}
\end{table}

\begin{table}
\centering
\renewcommand{\arraystretch}{1.5}
\renewcommand{\tabcolsep}{1.2mm}
    \caption{Performance on the proposed dataset. For fair comparison, all deep-learning based competing models are fine-tuned on the proposed dataset.}
    \label{tab:perform3}
    \scriptsize
    \centering
    \begin{tabular}{|c|cccccc|}

    \hline
    {\bfseries }  &{\bfseries CC$\uparrow$} &{\bfseries NSS$\uparrow$} &{\bfseries AUC-Borji$\uparrow$} &{\bfseries sAUC$\uparrow$} &{\bfseries SIM$\uparrow$} &{\bfseries KL$\downarrow$} \\
   \hline
   \hline

    \hline
    {{\bfseries GazeGAN}}  &{\bfseries 0.760} &{\bfseries 2.140} &{\bfseries 0.865} &{\bfseries 0.643} &{ 0.663} &{\bfseries 0.781}\\
    \hline

    \hline
    {{ SAM-VGG \cite{SAM}}}  &{0.753} &{2.134} &{0.859} &{0.612} &{\bfseries 0.668} &{0.831}\\
    \hline

    \hline
    {{ SAM-ResNet \cite{SAM}}} &{\bfseries 0.760} &{2.128} &{0.862} &{0.622} &{0.659} &{0.878}\\
    \hline

    \hline
    {{ ML-Net \cite{MLnet}}}  &{0.586} &{1.698} &{0.793} &{0.623} &{0.541} &{0.796}\\
    \hline

    \hline
    {{ OpenSALICON \cite{SALICON}}}  &{0.543} &{1.539} &{0.822} &{0.634} &{0.511} &{ 0.783}\\
    \hline

    \hline
    {{ SalGAN \cite{SalGAN}}}  &{0.561} &{1.524} &{0.820} &{0.633} &{0.489} &{ 0.864}\\
    \hline

    \hline
    {{ Sal-Net \cite{SalNet}}}  &{0.553} &{1.433} &{0.828} &{0.600} &{0.484} &{ 0.874}\\
    \hline

    \hline
    {{ GBVS \cite{GBVS}}}            &{0.521} &{1.341} &{0.821} &{0.585} &{0.468} &{ 0.879}\\
    \hline

    \hline
    {{ Itti\&Koch \cite{IttiKoch}}}  &{0.439} &{1.118} &{0.783} &{0.582} &{0.430} &{ 1.021}\\
    \hline

     \hline
    \end{tabular}
\end{table}

\begin{table}
\centering
\renewcommand{\arraystretch}{1.5}
\renewcommand{\tabcolsep}{.5mm}
    \caption{Performance on SALICON (old version) validation set \cite{SALICONDB}. For fair comparison, all competitors are trained from scratch.}
    \label{tab:perform1}
    \scriptsize
    \centering
    \begin{tabular}{|c|ccccccc|}

    \hline
    {\bfseries } &{\bfseries AUC-Judd$\uparrow$} &{\bfseries CC$\uparrow$} &{\bfseries NSS$\uparrow$} &{\bfseries AUC-Borji$\uparrow$} &{\bfseries sAUC$\uparrow$} &{\bfseries SIM$\uparrow$} &{\bfseries KL$\downarrow$} \\
   \hline
   \hline

    \hline
    {{ \bfseries GazeGAN}} &{\bfseries 0.891} &{\bfseries 0.808} &{\bfseries 2.914} &{\bfseries 0.878} &{\bfseries 0.743} &{\bfseries 0.764} &{\bfseries 0.496}\\
    \hline

     \hline
    {{ SAM-VGG \cite{SAM}}} &{0.879} &{0.756} &{2.900} &{0.850} &{0.712} &{0.722} &{0.545}\\
    \hline

    \hline
    {{ SAM-ResNet \cite{SAM}}} &{0.886} &{0.774} &{2.860} &{0.856} &{0.727} &{0.733} &{0.533}\\
    \hline

 \hline
    {{ OpenSALICON \cite{SALICON}}} &{0.886} &{0.748} &{2.823} &{0.833} &{0.726} &{0.720} &{0.516}\\
    \hline

 \hline
    {{ ML-Net \cite{MLnet}}} &{0.863} &{0.669} &{2.392} &{0.840} &{0.704} &{0.716} &{0.577}\\
    \hline

    \hline
    {{ SalGAN \cite{SalGAN}}} &{0.807} &{0.703} &{1.987} &{0.810} &{0.707} &{0.712} &{0.580}\\
    \hline

    \hline
    {{ Sal-Net \cite{SalNet}}} &{0.853} &{0.557} &{1.430} &{0.803} &{0.677} &{0.690} &{0.615}\\
    \hline

     \hline
    \end{tabular}
\end{table}

\subsection{Ablation analysis}
In this section, we evaluate the contribution of each component of the proposed model. We first compare the performance of GazeGAN when using different losses, as shown in Table \ref{tab:AblaLoss}. We find that the combination of pixel-level losses, histogram loss, and adversarial loss achieves superior performance over different evaluation metrics.

Next, we focus on the contributions of different modules of our model. For this purpose, we construct four different variations: $V_1$: the plain U-Net, $V_2$: the plain U-Net integrated with four residual blocks, $V3$: the modified $V2$ equipped with the CSC module, and $V_4$ is constructed by appending the local generator to $V_3$. Table \ref{tab:AblaModule} shows the ablation analysis results on LSUN'17 validation set. We can see that every module contributes to the final performance.
We provide more ablation study results on SALICON (previously released), MIT1003, and the proposed dataset in the supplementary material.

\begin{figure*}[htbp!]
\center
\includegraphics[scale=0.185]{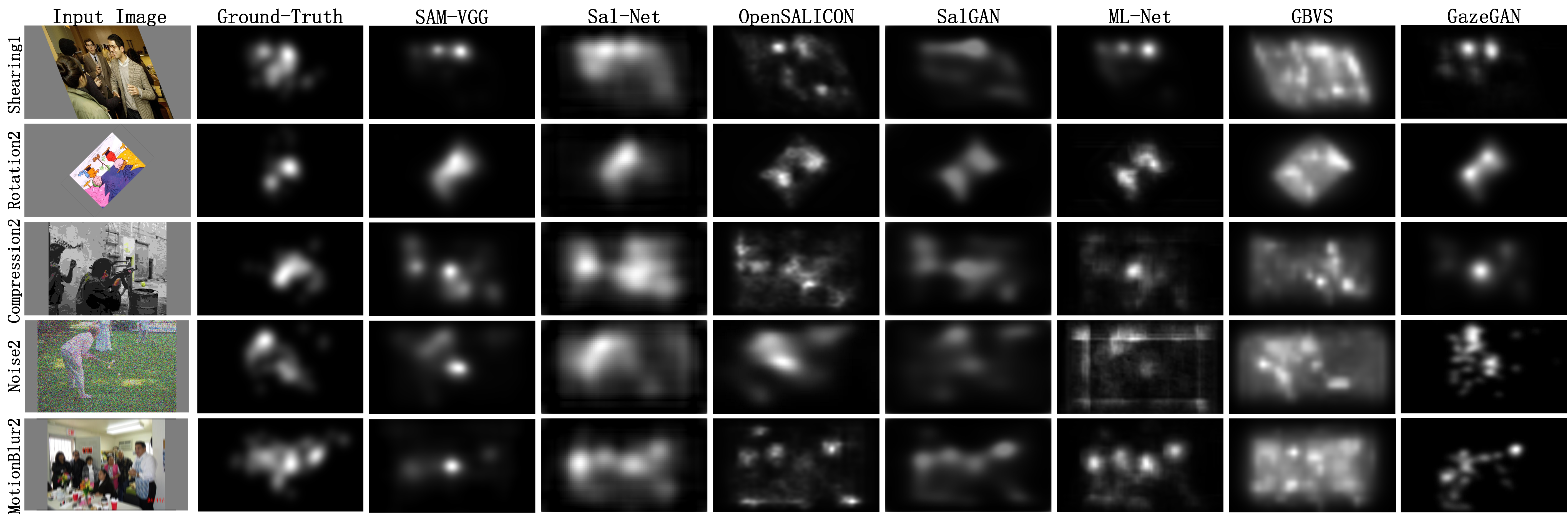}%0.2
\caption{ Qualitative results on various transformed scenes of the proposed dataset.}
\label{ourrs}
\end{figure*}

\begin{figure}[htbp!]
\center
\includegraphics[scale=0.35]{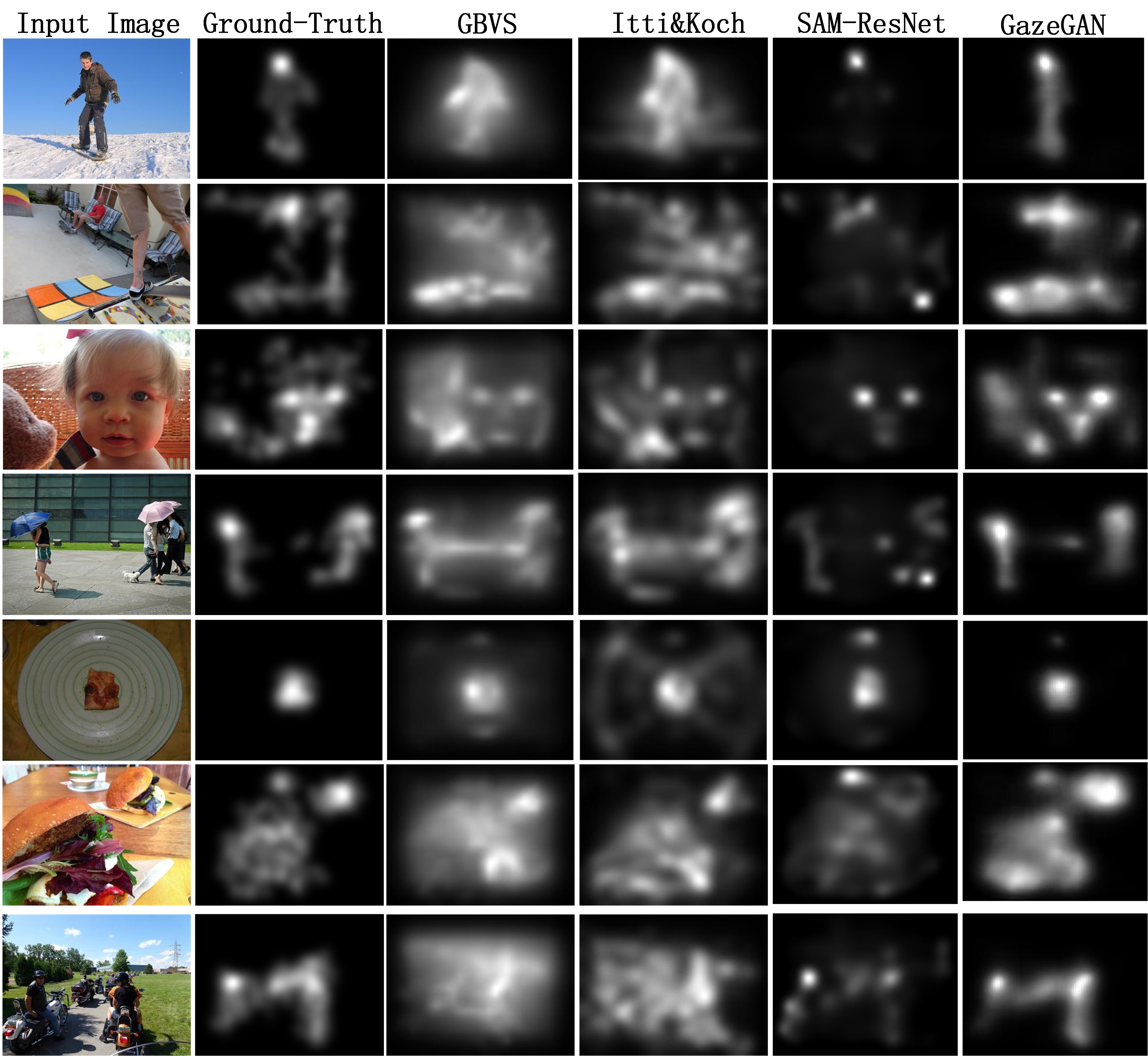}
\caption{ Qualitative results on normal stimuli of SALICON \cite{SALICONDB}.}
\label{saliconrs}
\end{figure}

\begin{figure}[htbp!]
\center
\includegraphics[scale=0.21]{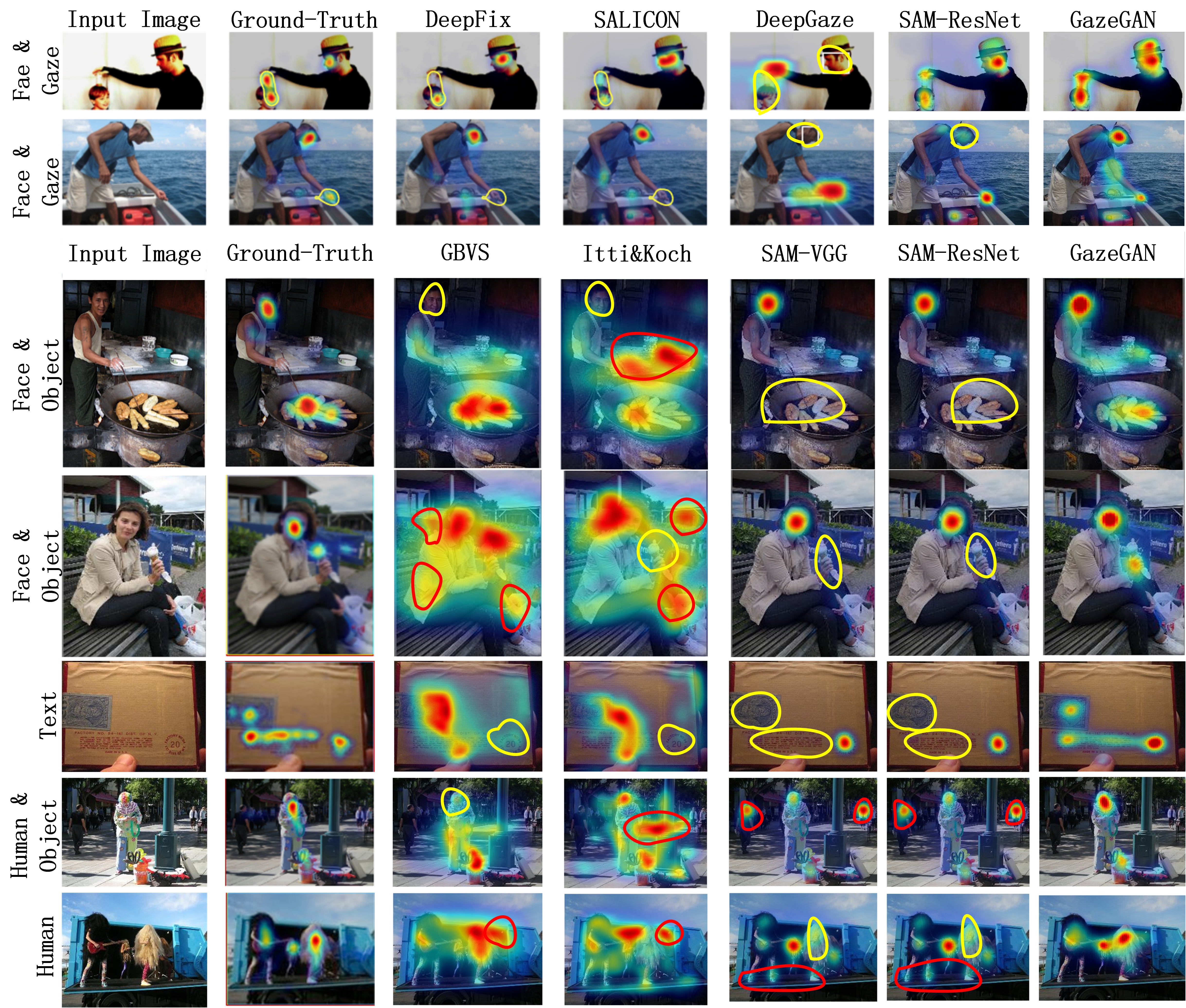}
\caption{ Visualization on normal MIT300 dataset \cite{mit-saliency-benchmark}. Yellow (red) polygons represent missing regions (wrongly detected regions).}
\label{mit300rs}
\end{figure}

\subsection{Comparison with the state-of-the-art}
We first quantitatively compare GazeGAN with state-of-the-art models on SALICON (old version), MIT1003, LSUN'17 (SALICON-2017-version), and the proposed dataset. Experimental results are reported in Tables \ref{Test2017}-\ref{tab:perform1}.
GazeGAN achieves top-ranked performance on the SALICON (old version) validation set and proposed dataset over different evaluation metrics. It also obtains competitive performance on the MIT1003 and LSUN'17 datasets.

The qualitative results are shown in Figs.~\ref{ourrs}-\ref{mit300rs}. We notice that, GazeGAN generates accurate results for various transformed scenes, as in Fig.~\ref{ourrs}. Besides, on normal stimuli in Fig.~\ref{saliconrs} and Fig.~\ref{mit300rs}, GazeGAN performs well, even for challenging scenes containing multiple faces, gazed-upon objects and text, as in Fig.~\ref{mit300rs}.

\subsection{Finer-grained comparison on transformed dataset}
\begin{figure*}%[!htb]
%\vspace{-0.1cm}
\centering
\subfigure[sAUC]{\label{fig:edge-a}\includegraphics[height=0.42\linewidth]{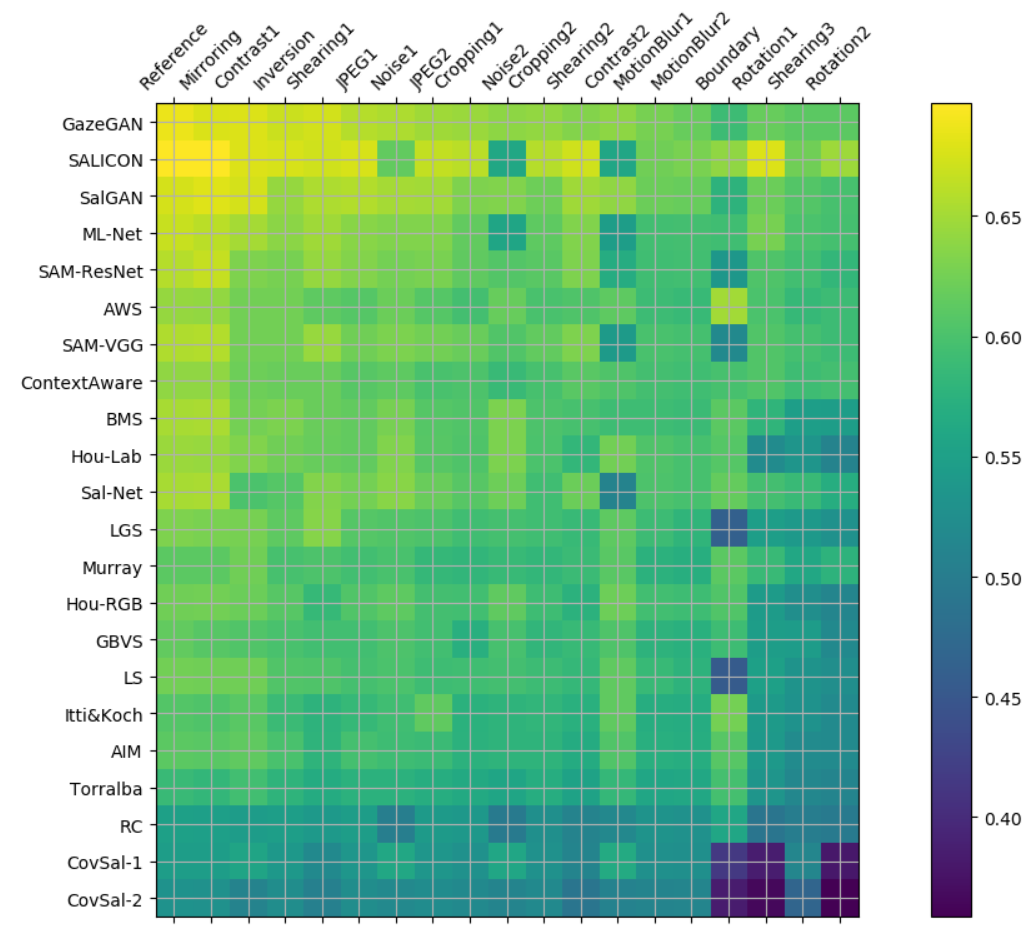}}
\subfigure[NSS]{\label{fig:edge-c}\includegraphics[height=0.42\linewidth]{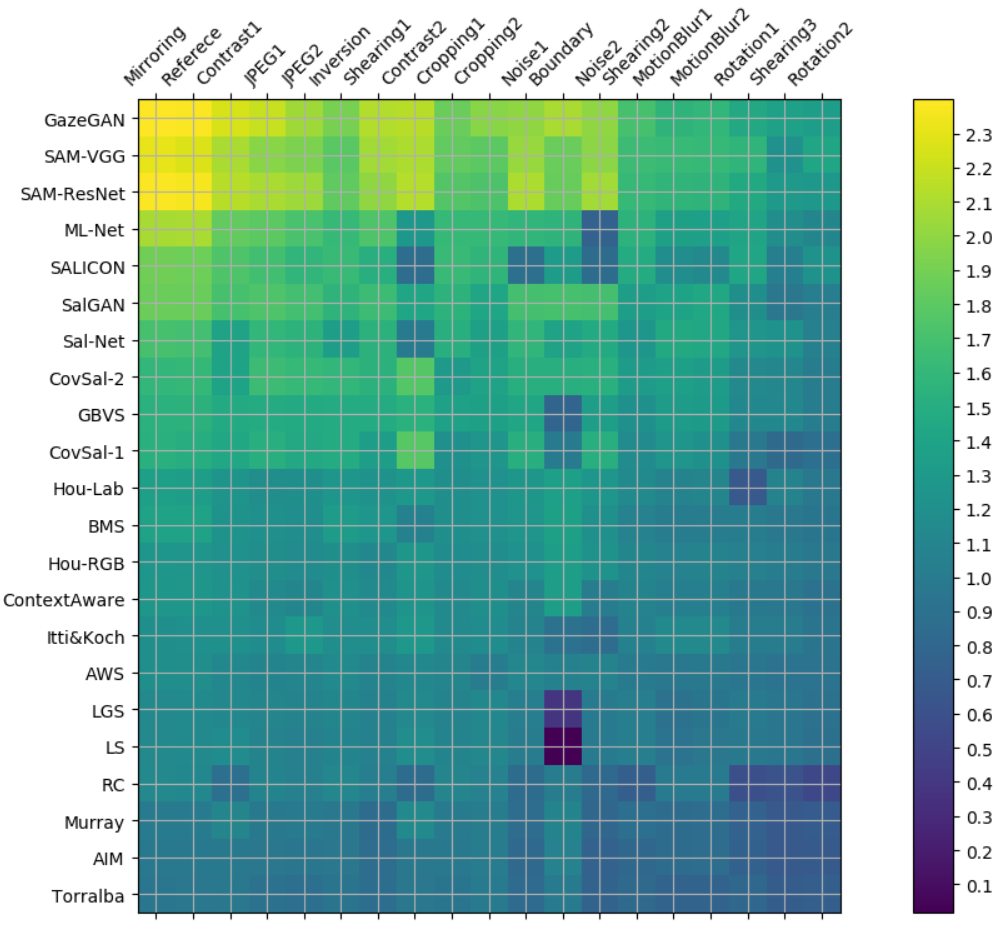}}
\caption{ Fine-grained performance comparison of state-of-the-art saliency models on different transformations of the proposed dataset. The horizontal axis represents different transformation types which are ranked by average performance over 22 saliency models. The vertical axis represents different saliency models which are ranked by average performance over 19 transformations. This comparison provides a benchmark for saliency models on transformed stimuli.}
\label{fgperformances}
\end{figure*}

As shown in Fig.~\ref{fgperformances},
we further provide the fine-grained comparison of 22 existing saliency models on each transformation type of the proposed dataset.

For comprehensive comparison, we select 15 early saliency models based on hand-crafted features, $i.e.$ Itti$\&$Koch \cite{IttiKoch}, GBVS \cite{GBVS}, Torralba \cite{Torralba}, CovSal \cite{Covsal} (CovSal-1 utilizes covariance feature and CovSal-2 utilizes both of covariance and mean features), AIM \cite{AIM}, Hou \cite{ImageSig} (Hou-Lab and Hou-RGB adopt Lab and RGB color spaces respectively), LS \cite{LSLGS}, LGS \cite{LSLGS}, BMS \cite{BMS}, RC \cite{RC}, Murray \cite{Murray}, AWS \cite{AWS} and ContextAware \cite{ContextAware}. We also select 7 deep saliency models, $i.e.$ GazeGAN, ML-Net \cite{MLnet}, SalGAN \cite{SalGAN}, OpenSALICON \cite{SALICON}, Sal-Net \cite{SalNet}, SAM-ResNet \cite{SAM} and SAM-VGG \cite{SAM}.

We observe the following points from Fig. \ref{fgperformances}:\footnote{We provide more results under CC and KL metrics in the supplement.}

\begin{itemize}
    \item {$\bf Challenging$ $\bf Transformations$:} Rotation2, Shearing3, Noise2 and Contrast2 are the most challenging transformations for saliency models. Most saliency models underperform on these transformations. Rotation2 and Shearing3 impose severe geometrical transformations, while Noise2 and Contrast2 include high level spatial perturbations. The former changes the spatial structure of image, while the latter alters intensities and local contrast. Recall that Rotation2 and Shearing3 also have severe impacts on human gaze.
    \item {$\bf Outliers$:} LS and LGS fail on Boundary. Sal-Net fails on Contrast2. ML-Net and OpenSALICON fail on Noise2 and Contrast2. CovSal-1 and CovSal-2 fail on sAUC metric, especially on Rotation and Boundary, because the CovSal model overemphasizes center-bias which is penalized by the sAUC metric.
    \item {$\bf Deep$ $\bf Models$ $ vs.$ $\bf Early$ $\bf Models$:} Deep saliency models obtain higher performances compared to the early models based on hand-crafted features. GazeGAN achieves top-ranked average performance over different metrics. Besides, GazeGAN is robust to various types of transformation without obvious failures.
\end{itemize}

\subsection{Discussion on the robustness of GazeGAN}

\begin{figure*}[htbp!]
%\vspace{-0.1cm}
\center
\includegraphics[scale=0.34]{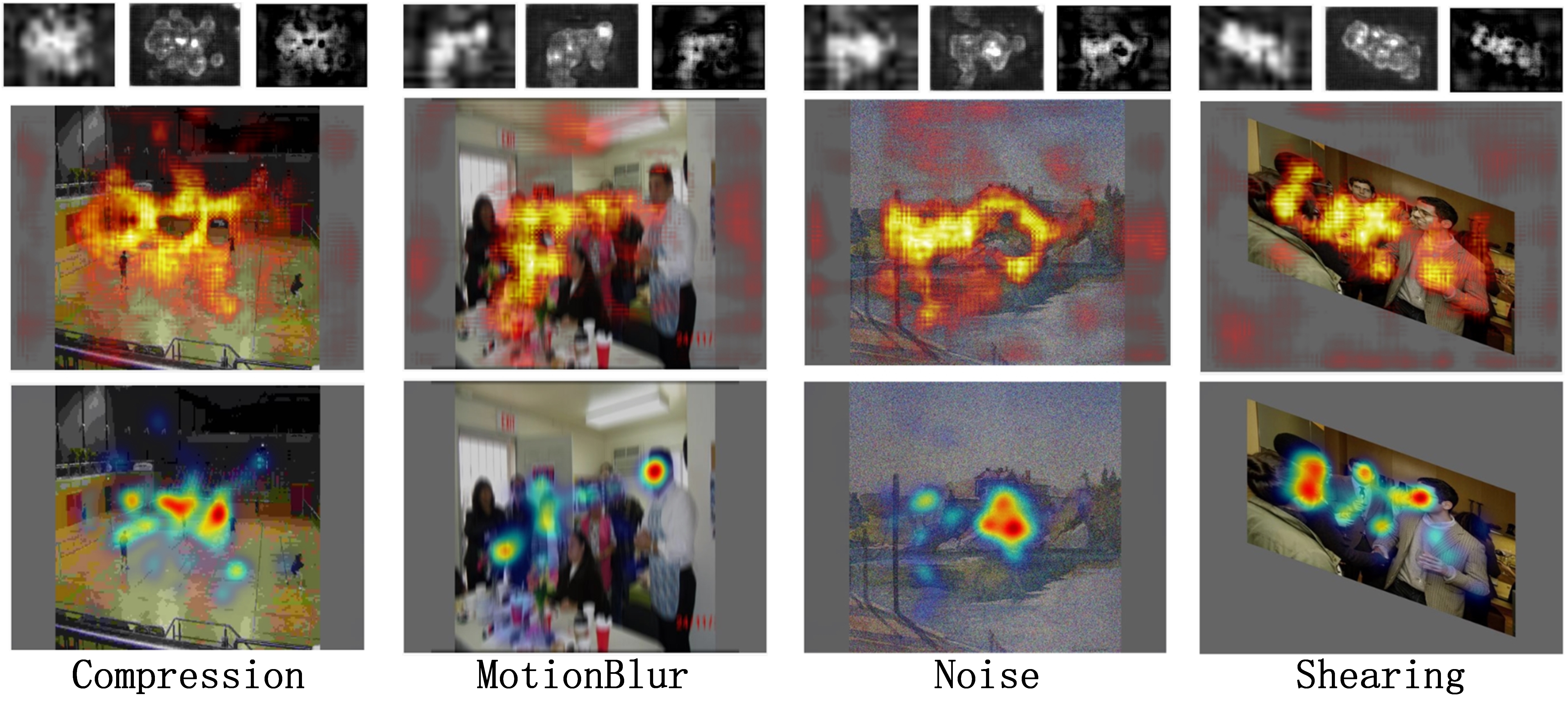}
\caption{Visualizations of the proposed CSC module. The $1_{st}$ row represents the \textit{surround} feature maps (from decoder layer1), \textit{center} feature maps (from decoder layer5), and the \textit{difference} maps of \textit{surround} and \textit{center}, respectively. The $2_{nd}$ row reflects the wrong predictions of \textit{surround} feature maps caused by trivial artifacts, while the $3_{rd}$ row reflects the final predictions processed by CSC module that focus on semantic salient regions.
The feature maps of the $1_{st}$ row are normalized by average pooling in the channel direction, then we use bilinear interpolation to upsample the feature maps to have the same resolution as the input image for better observation, as shown in the $2_{nd}$ and $3_{rd}$ rows.}
\label{cscvis}
\end{figure*}

As indicated in Fig.~\ref{ourrs}, Fig.~\ref{fgperformances} and Table~\ref{tab:perform3}, the proposed GazeGAN achieves better robustness against various transformations. In this section, we discuss the robustness of the proposed model from different perspectives.

\begin{itemize}
    \item {$\bf Advantages$ $\bf of$ $\bf CSC$:} Our proposed CSC module has two advantages.
It mitigates the trivial artifacts, and highlights semantic salient information, as shown in Fig.~\ref{cscvis}. For example, in the $1_{st}$ column of Fig.~\ref{cscvis}, we notice that the compression artifacts cause wrong predictions in the \textit{surround} feature maps, and we want to mitigate the impacts of these trivial artifacts. Besides, despite the \textit{surround} feature maps can detect semantic salient regions (\eg ``pedestrians''), the responses of semantic salient regions are not strong enough. Thus we want to further emphasize the responses of these semantic salient regions. We can see that the final output processed by CSC module concentrates on semantic salient regions, while ignoring the trivial artifacts.
\item {$\bf Model$ $\bf Nonlinearity$:} Second, CSC module improves the nonlinearity of the proposed deep model. Specifically, each individual CSC module contains three $1\times1$ convolution layers and one transposed convolution layer. We append a nonlinear ReLU activation after each convolution layer. Besides, we utilize eight CSC modules in the proposed GazeGAN architecture in total, that are $4\times8=32$ nonlinear activations. According to \cite{robust2,robust3}, the higher model nonlinearity increases the representational ability of deep neural network, demonstrating better robustness against transformations.
\item {$\bf Multiscale$ $\bf Network$ $\bf Architecture$:} Hendrycks~\etal~\cite{robust1} pointed that multiscale architectures achieve better robustness by propagating features across different scales at each layer rather than slowly gaining a global representation of the input as in traditional CNNs. GazeGAN utilizes both skip-connections, CSC connections, and local-gloabl GAN architectures. Both of these factors adequately leverage multiscale features.
\item {$\bf Hybrid$ $\bf Adversarial$ $\bf Training$:} Hybrid adversarial training is a defense strategy for improving robustness of deep CNN models against adversarial attacks \cite{robust3}. This method utilizes an ensemble of original images and the adversarial examples to train the deep models.
    Adversarial examples are the manually generated images by adding some slight perturbations to original images \cite{robust3}. In fact, the proposed {\emph {valid}} data augmentation strategy provides a similar solution, which is adopting the examples corrupted by an ensemble of several transformations to train the deep CNNs. This hybrid adversarial training strategy is currently the most effective method to improve model robustness, and prevents overfitting to a specific transformation type \cite{robust3}.
\end{itemize}

\section{Conclusion}
In this article, we introduce a new eye-tracking dataset containing several common image transformations. Based on our analyses of eye-movement data, we propose a {\emph {valid}} data augmentation strategy using some label-preserving transformations for boosting deep-learning based saliency models. Besides, we propose a new model called GazeGAN integrated with a novel center-surround connection module that mitigates trivial artifacts while emphasizing semantic salient regions, demonstrating better robustness against various transformations. GazeGAN achieves the best results on the transformed dataset, and obtains competitive performance on normal distortion-free benchmark datasets.
We share our dataset and code with the community at \url{https://github.com/CZHQuality/Sal-CFS-GAN}, where we provide both Pytorch and Tensorflow versions of the code. Our repository provides a flexible interface for users to integrate their own architectures and to promote research on improving the robustness of saliency models over non-canonical stimuli.

\ifCLASSOPTIONcaptionsoff
  \newpage
\fi

%\bibliographystyle{IEEEtran}      %LaTex Class文件, IEEEtran为给定模板格式定义文件名
%\bibliography{egbib}
{\small
\bibliographystyle{IEEEtran}
\bibliography{egbib}
}

% \begin{IEEEbiographynophoto}{Zhaohui Che}
\begin{IEEEbiography}[{\includegraphics[width=1in,height=1.25in,clip,keepaspectratio]{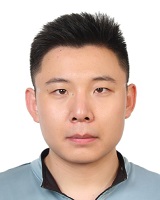}}]{Zhaohui Che}
received the B.E. degree from School of Electronic Engineering, Xidian University, Xi'an, China, in 2015. He is currently working toward the Ph.D. degree at the Institute of Image Communication and Network Engineering, Shanghai Jiao Tong University, Shanghai, China. His research interests include visual attention, perceptual quality assessment, deep learning, and adversarial attack and defense. From 2018 to 2019, he was a Visiting Student at the Ecole Polytechnique de l'Universite de Nantes, Nantes, France.
He won the Grand Prize of the ICME 2018 Grand Challenge on ``Salient360$!$'' for visual attention modeling for panoramic content. He was a co-organizer of the Grand Challenge ``Saliency4ASD'' at IEEE ICME 2019.
\end{IEEEbiography}

%\newpage

% \begin{IEEEbiographynophoto}{Ali Borji}
\begin{IEEEbiography}[{\includegraphics[width=1in,height=1.25in,clip,keepaspectratio]{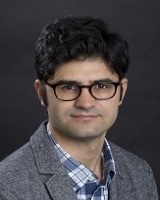}}]{Ali Borji}
received the B.S. degree in computer engineering from the Petroleum University of Technology, Tehran, Iran, in 2001, the M.S. degree in computer engineering from Shiraz University, Shiraz, Iran, in 2004, and the Ph.D. degree in cognitive neuro-sciences from the Institute for Studies in Fundamental Sciences, Tehran, Iran, in 2009. He spent four years as a Post-Doctoral Scholar with iLab, University of Southern California, from 2010 to 2014. He is currently a senior research scientist at MarkableAI Inc, Brooklyn, NY
11201, USA. His research interests include visual attention, active learning, object and scene recognition, and cognitive and computational neuro-sciences. He has published more than 150 academic papers, including IEEE Transactions on Pattern Analysis and Machine Intelligence, IEEE Transactions on Image Processing, IEEE International Conference on Computer Vision and Pattern Recognition, IEEE International Conference on Computer Vision, and European Conference on Computer Vision.
\end{IEEEbiography}

%\begin{IEEEbiographynophoto}{Guangtao Zhai}
\begin{IEEEbiography}[{\includegraphics[width=1in,height=1.25in,clip,keepaspectratio]{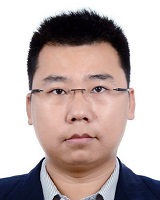}}]{Guangtao Zhai}
received the B.E. and M.E. degrees from Shandong University, Shandong, China, in 2001 and 2004, respectively, and the Ph.D. degree from Shanghai Jiao Tong University, Shanghai, China, in 2009. He is currently a Research Professor with the Institute of Image Communication and Information Processing, Shanghai Jiao Tong University. From 2008 to 2009, he was a Visiting Student at the Department of Electrical and Computer Engineering, McMaster University, Hamilton, ON, Canada, where he was a Post-Doctoral Fellow from 2010 to 2012. From 2012 to 2013, he was a Humboldt Research Fellow with the Institute of Multimedia Communication and Signal Processing, Friedrich Alexander University of Erlangen-Nuremberg, Erlangen, Germany. His research interests include multimedia signal processing and perceptual signal processing. Prof. Zhai was the recipient of the Award of National Excellent Ph.D. thesis from the Ministry of Education of China in 2012.
\end{IEEEbiography}

% \begin{IEEEbiographynophoto}{Xiongkuo Min}
\begin{IEEEbiography}[{\includegraphics[width=1in,height=1.25in,clip,keepaspectratio]{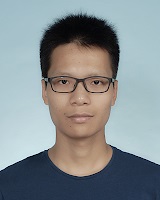}}]{Xiongkuo Min}
received the B.E. degree from Wuhan University, Wuhan, China, in 2013. He received the Ph.D. degree at the Institute of Image Communication and Network Engineering, Shanghai Jiao Tong University, Shanghai, China, in 2018. From 2016 to 2017, he was a Visiting Student at the Department of Electrical and Computer Engineering, University of Waterloo, Waterloo, ON, Canada. His research interests include image quality assessment, visual attention modeling, and perceptual signal processing. Mr. Min was the recipient of the Best Student Paper Award of ICME 2016.
\end{IEEEbiography}

% \begin{IEEEbiographynophoto}{Guodong Guo}
\begin{IEEEbiography}[{\includegraphics[width=1in,height=1.25in,clip,keepaspectratio]{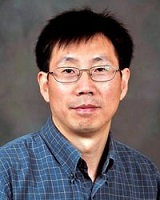}}]{Guodong Guo}
received the B.E. degree in automation from the Tsinghua University, Beijing, China, the PhD degree in pattern recognition and intelligent control from the Chinese Academy of Sciences, Beijing, China, and the PhD degree in computer science from the University of Wisconsin-Madison, Madison, WI, in 2006. He is an associate professor with the Department of Computer Science and Electrical Engineering, West Virginia University (WVU), Morgantown, WV. In the past, he visited and worked in several places, including INRIA, Sophia Antipolis, France; Ritsumeikan University, Kyoto, Japan; Microsoft Research, Beijing, China; and North Carolina Central University. He authored a book, ``Face, Expression, and Iris Recognition Using Learning-Based Approaches'' (2008), co-edited a book, ``Support Vector Machines Applications'' (2014), and published over 60 technical papers. His research interests include computer vision, machine learning, and multimedia. He received the North Carolina State Award for Excellence in Innovation in 2008, Outstanding Researcher in 2013-2014 at the CEMR, WVU, and New Researcher of the Year in 2010-2011 at the CEMR, WVU. He was selected the “People's Hero of the Week” by BSJB under Minority Media and Telecommunications Council (MMTC) on July 29, 2013. Two of his papers were selected as ``The Best of FG'13'' and ``The Best of FG'15'', respectively. He is a senior member of the IEEE. He joined the Institute of Deep Learning, Baidu Research, in 2018.
\end{IEEEbiography}

% \begin{IEEEbiographynophoto}{Patrick Le Callet}
\begin{IEEEbiography}[{\includegraphics[width=1in,height=1.25in,clip,keepaspectratio]{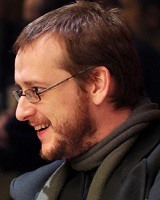}}]{Patrick Le Callet}
was born in 1970. He received both the M.Sc. and Ph.D. degrees in image processing from Ecole Polytechnique de l'Universite de Nantes, Nantes, France. He was also a student at the Ecole Normale Superieure de Cachan where he sat the Aggregation (credentialing exam) in electronics of the French National Education. He was an Assistant Professor from 1997 to 1999 and a Full Time Lecturer from 1999 to 2003 with the Department of Electrical Engineering, Technical Institute of the University of Nantes. Since 2003, he has been teaching with the Departments of Electrical Engineering and Computer Science, Engineering School, Ecole Polytechnique de l'Universite de Nantes where he is currently a Full Professor. He is mostly engaged in research dealing with the application of human vision modeling in image and video processing. He is a co-author of more than 250 publications and communications and co-inventor of 16 international patents. His current research interests include quality of experience assessment, visual attention modeling and applications, perceptual video coding, and immersive media processing. He serves or has served as an Associate Editor or Guest Editor for several journals including IEEE Transactions on Image Processing, IEEE Journal of Selected Topics in Signal Processing, IEEE Transactions on Circuits and Systemsfor Video Technology, and Springer Eurasip Journal on Image and Video Processing. He is the IEEE fellow.
\end{IEEEbiography}


%\vfill


\end{document}